\documentclass[runningheads]{llncs}

\usepackage{eccv}

\usepackage{eccvabbrv}

\usepackage{graphicx}
\usepackage{booktabs}
\definecolor{Rednew}{RGB}{128,0,0}

\usepackage{multirow}
\usepackage{wrapfig}
\usepackage{epsfig}
\usepackage{adjustbox}
\usepackage{amsmath}
\usepackage{amssymb}
\usepackage{marvosym}
\usepackage{threeparttable}
\usepackage{algorithm}
\usepackage{algpseudocode}
\usepackage{caption}
\usepackage{colortbl}
\usepackage{listings}
\usepackage{pifont}

\usepackage{array}   
\usepackage{bm}       

\usepackage[accsupp]{axessibility}  

\usepackage{hyperref}

\usepackage{orcidlink}

\begin{document}

\title{Generalized Biomedicine Discovery} 

\titlerunning{Generalized Biomedicine Discovery}

\author{Luyao Tang \inst{1}\orcidlink{0009-0007-2447-3194} \and
Yingkai Yang\inst{2}\orcidlink{0009-0004-0681-4993} \and
Hanqi Chen\inst{2}\orcidlink{0009-0009-8876-5289} \and
Jiewei Zheng\inst{3}\orcidlink{0009-0005-1872-540X} \and
\\Chaoqi Chen\inst{2}\orcidlink{0000-0002-0052-3819}
\and
Cheng Chen$^{\dagger}$\inst{1}\orcidlink{0000-0002-6040-6833}}

\authorrunning{L.~Tang \and Y.~Yang et al.}

\institute{The University of Hong Kong, Hong Kong SAR, China \and
Shenzhen University, Shenzhen, China  \and
Xiamen University, Xiamen, China \\
\email{lytang1999@gmail.com}, \email{cchen@eee.hku.hk}
}

\maketitle

\begingroup
\renewcommand\thefootnote{}
\footnotetext{$^{\dagger}$ Corresponding author.}
\endgroup

\begin{abstract}
In real-world clinical practice, medical images face open-world shifts: (i) long-tailed rare diseases, (ii) subtle lesions dominated by normal anatomy, and (iii) hierarchical taxonomies. Yet most open-world paradigms assume flat, balanced label spaces, leaving these biomedical demands unresolved. We introduce Generalized Biomedicine Discovery (\textbf{GBD}) and a unified benchmark spanning long-tail, anomaly, and taxonomy-aware discovery. Our key insight is that dominant known patterns form a visual manifold that masks subtle novelty. Inspired by expert diagnosis, we propose \textbf{SCAN} (Surprise-evoked Complementary AccommodatioN), which follows a cognition-inspired perceptual progression: it applies \textit{predictive suppression} to filter expected norms, triggers \textit{surprise-evoked salience} to highlight unexpected deviations, and performs \textit{complementary accommodation} to integrate these shifts into global representations. Extensive experiments show that SCAN improves novel concept discovery while generally preserving established clinical knowledge, and it plugs into existing architectures to better navigate the known--unknown trade-off in medical imaging. Code is available at \href{https://github.com/lytang63/generalized-biomedicine-discovery}{\textcolor{purple}{github.com/lytang63/generalized-biomedicine-discovery}}.
\end{abstract}

\section{Introduction}
\label{sec:intro}

Medical imaging systems are increasingly deployed in \emph{open-world} clinical environments~\cite{zhang2025generalist,yang2025medical}, where the test-time concept space is neither closed nor balanced~\cite{yang2024limits}. In clinical practice, medical images exhibit three characteristic open-world signatures: (i) \textbf{long-tailed} distributions~\cite{yang2022proco,zhang2023deep,zheng2024large,holste2022long} where clinically critical rare diseases are sparsely observed, (ii) the \textbf{visual dominance} of normal anatomy~\cite{thirion2000statistical,bushberg2011essential,feeman2010mathematics,he2023transnuseg} that overwhelms subtle lesions, and (iii) \textbf{hierarchical} taxonomies~\cite{dimitrovski2011hierarchical,chen2020deep,yu2025hierarchical} where novel concepts emerge as fine-grained siblings within established diagnostic families. Yet, most open-world paradigms and benchmarks still assume flat and relatively balanced concept spaces, leaving these authentic biomedical needs largely unresolved~\cite{feng2025neighbor,das2025medgcd}.

\begin{figure}[tb]
  \centering
  \includegraphics[width=0.98\textwidth]{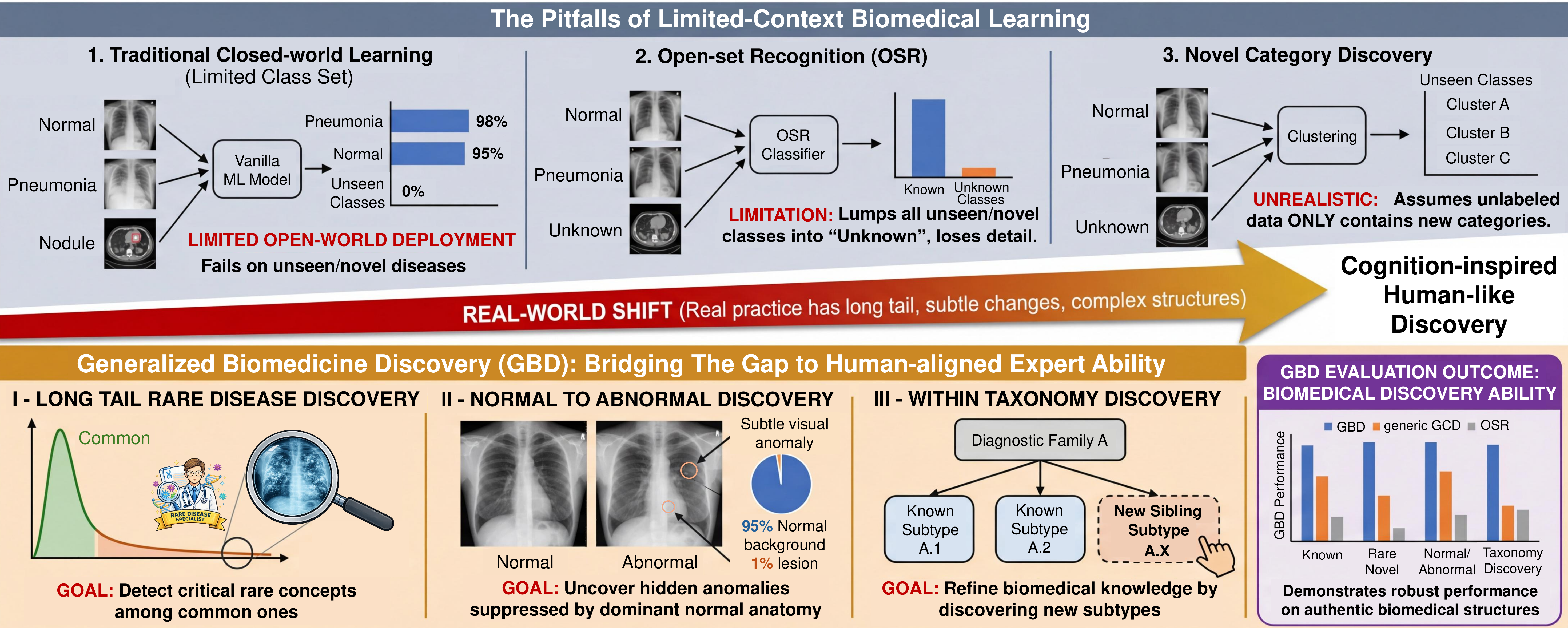}
\caption{\textbf{GBD in practice.} Real biomedical data shift beyond the closed world: standard classification fails on unseen diseases, OSR merges heterogeneous novelties into one ``unknown'', and conventional discovery often assumes unlabeled data contains only new categories. \textbf{GBD} formalizes three realistic discovery regimes: (I) long-tail rare disease discovery among common cases, (II) normal-to-abnormal discovery of subtle lesions dominated by normal anatomy, and (III) within-taxonomy discovery that uncovers new sibling subtypes to refine diagnostic hierarchies.}
  \label{fig:clarify}
\end{figure}

As illustrated in~\cref{fig:clarify}, conventional closed-set learning assumes a static class vocabulary, limiting its use in dynamic clinical environments. Open-Set Recognition (OSR)~\cite{saito2021openmatch,yoshihashi2019classification,ssb,geng2020recent,scheirer2012toward} identifies unseen classes but lumps them into a single ``unknown'' category, while Novel Category Discovery (NCD)~\cite{openmix,incd,ncd} clusters unknowns under the unrealistic assumption that unlabeled data contains \emph{only} novel classes. Generalized Category Discovery (GCD)~\cite{gcd,he2025category,gpc,dccl,legogcd,tangbures,sptnet,Bian2026Foundation,tang2025dissecting} instead jointly classifies known and discovers novel classes in mixed unlabeled data. However, these paradigms do not capture biomedical data, where class imbalance, spatially dominant normal anatomy, and structured clinical taxonomies jointly obscure novel concepts~\cite{zheng2024large,zhang2023deep,thirion2000statistical,lester1999survey,yao2023adversarial}. Consequently, methods validated on natural-image GCD often fail to translate to clinical settings.

This mismatch motivates an expert-aligned target: clinicians discover new concepts \emph{within} established knowledge by recognizing expected patterns, detecting subtle deviations, and situating new subtypes within diagnostic families~\cite{abu2017medical,shortliffe1986medical,gilhooly1990cognitive}. Existing open-world paradigms~\cite{tang2025ocrt,das2025medgcd,feng2025neighbor,gcd,ncd,chen2022mix} rarely evaluate this capability, especially under long-tail rarity and hierarchical taxonomies~\cite{lester1999survey,yao2023adversarial}.

To bridge this gap, we introduce \textbf{Generalized Biomedicine Discovery (GBD)}, a realistic open-world paradigm and unified medical-imaging benchmark with three clinically grounded settings: (1) \textit{Long-tail Rare Disease Discovery}, (2) \textit{Normal-to-Abnormal Discovery}, and (3) \textit{Within-Taxonomy Discovery}. Beyond these structural challenges, GBD exposes a critical visual problem: dominant known features (\emph{e.g.}, widespread normal anatomy) suppress subtle, localized novel signals during global representation learning. This bias attenuates the residual evidence needed to discover new diseases, motivating a mechanism that counters visual suppression without prior knowledge of the anomalies.

Inspired by diagnostic cognition~\cite{gilhooly1990cognitive,van2013cognitive,croskerry2023challenge}, we propose \textbf{SCAN} (Surprise-evoked Complementary AccommodatioN), a plug-and-play layer operating on a vision backbone's class and patch tokens~\cite{dino}. SCAN instantiates this progression in three operations: (i) \textbf{predictive suppression} filters anticipated patterns under predictive coding~\cite{rao1999predictive}; (ii) \textbf{surprise-evoked salience} uses residual energy to amplify unexpected, localized deviations~\cite{itti2009bayesian}; and (iii) \textbf{complementary accommodation} integrates the isolated novelty into the global representation while preserving established structure~\cite{mcclelland1995there}.

\begin{itemize}
\item We introduce \textbf{GBD}, a clinically grounded open-world paradigm and unified benchmark for biomedicine, formulating three grounded settings: Long-tail Rare Disease, Normal-to-Abnormal, and Within-Taxonomy Discovery.
\item We propose \textbf{SCAN}, a plug-and-play cognitive vision layer that mitigates dominant-feature suppression through predictive filtering, surprise amplification, and complementary feature accommodation.
\item Extensive experiments demonstrate that SCAN generally improves conventional GCD methods on the GBD benchmark, with particularly strong gains for SelEx in isolating subtle anomalies and discovering long-tailed rare diseases, while some baseline- and subset-specific trade-offs remain.
\end{itemize}

\section{Related Work}
\label{sec:related}

\paragraph{\textbf{Open-world Learning.}}
Open-world learning aims to identify and discover novel categories beyond closed-set assumptions. Open-set recognition~\cite{saito2021openmatch,yoshihashi2019classification,ssb,geng2020recent,he2026flowcomposer,he2026harnessing,he2026task,chen2023activate} detects unseen classes but groups them into a single "unknown" label, lacking fine-grained discrimination. Novel category discovery~\cite{openmix,incd,ncd} clusters unknowns yet unrealistically assumes unlabeled data contains only novel classes. Generalized category discovery~\cite{gcd} advances this by jointly classifying knowns and discovering novelties in mixed unlabeled data, with methods split into parametric~\cite{simgcd, legogcd, ugcd} and non-parametric frameworks~\cite{pim, gpc, cms, selex}. GCD commonly refines representations through contrastive learning~\cite{simclr,he2020momentum,liu2025hyperbolic,liu2025debgcd,tangcoge}, but often assumes separable pre-trained features. Recent variants extend GCD to continual open-world shifts~\cite{fengprism,feng2026beyond,feng2026seeing}, while medical approaches adapt discovery to clinical images~\cite{feng2025neighbor,feng2026leveraging}. Yet these efforts largely retain flat category spaces, leaving biomedical class imbalance, dominant normal anatomy, and clinical hierarchies underexplored. We instead target dominant-feature suppression with a cognition-aligned framework tailored to biomedical discovery.

\paragraph{\textbf{Cognitive Sciences in Machine Learning.}}
Cognitive sciences principles have increasingly informed machine learning, leveraging human perceptual and reasoning mechanisms for more robust generalization~\cite{tang2025ocrt,thagard2013cognitive,blasi2022over,chen1982topological}. Prior work draws on structured cognitive processes (\emph{e.g.}, perceptual grouping, inductive reasoning)~\cite{yang2024language} to design structured priors~\cite{sung2015data,wen2022mvs,roessle2022dense} and factorized representations~\cite{song2023flow,liang2023factorized}, enhancing model robustness across tasks~\cite{muennighoff2023crosslingual,zhang2023federated}. Key cognitive theories: predictive coding~\cite{rao1999predictive} (suppressing anticipated visual features), surprise-driven salience~\cite{itti2009bayesian} (amplifying anomalous signals), and complementary learning systems~\cite{mcclelland1995there} have been explored in isolation for visual processing, yet remain underexploited for biomedical diagnostic cognition~\cite{gilhooly1990cognitive, van2013cognitive, croskerry2023challenge}. We integrate these theories into a unified framework that mirrors clinical diagnostic reasoning, filling this critical gap for biomedical open-world learning.


\section{Problem Setting}
\label{sec:setting}

\subsection{Task Backbone: Generalized Biomedicine Discovery (GBD)}
For a given dataset, let $\mathcal{X}$ and $\mathcal{Y}$ denote the image and label spaces. We are provided with a labeled subset
$\mathcal{D}_l=\{(\mathbf{x}_i^l,y_i^l)\}\subset \mathcal{X}\times \mathcal{Y}_l$
and an unlabeled subset
$\mathcal{D}_u=\{(\mathbf{x}_i^u,y_i^u)\}\subset \mathcal{X}\times \mathcal{Y}_u$. $y_i^u$ is used only for evaluation.
Only \emph{known} classes are labeled in $\mathcal{D}_l$, while $\mathcal{D}_u$ contains a mixture of known and \emph{novel} classes:
$
\mathcal{Y}_l=\mathcal{C}_{\mathrm{known}}$  and $
\mathcal{Y}_u=\mathcal{C}_{\mathrm{known}}\cup \mathcal{C}_{\mathrm{novel}}.
$
The objective of GBD is to assign each unlabeled sample to either a known class or a newly discovered novel class. 

\subsection{Setting I: Long-tail Rare Disease Discovery}
This setting instantiates discovery under the \emph{long-tail statistics} prevalent in biomedicine. Let $\mathcal{C}$ denote the set of leaf-level diagnostic categories, and $n(c)$ be the empirical frequency of class $c\in\mathcal{C}$ in the training pool. We define a head--tail split controlled by a parameter $\alpha\in(0,1)$:
\begin{equation}
\mathcal{C}_{\mathrm{known}}=\mathrm{Head}_{\alpha}(\mathcal{C}) \;=\; \arg\max_{\substack{\mathcal{S}\subseteq \mathcal{C}\\ |\mathcal{S}|=\lceil \alpha|\mathcal{C}|\rceil}}\;\sum_{c\in\mathcal{S}} n(c),\qquad
\mathcal{C}_{\mathrm{novel}}=\mathcal{C}\setminus \mathcal{C}_{\mathrm{known}},
\end{equation}
where ties in empirical frequencies are broken lexicographically. Thus, $\mathcal{D}_l$ contains labels only from $\mathcal{C}_{\mathrm{known}}$, while $\mathcal{D}_u$ mixes samples from $\mathcal{C}_{\mathrm{known}}\cup\mathcal{C}_{\mathrm{novel}}$. This design mirrors the \underline{epidemiological reality} where annotations focus on frequent conditions, forcing the model to discover \emph{rare but clinically critical} tail diseases.
We generally use $\alpha=0.5$, except for classes explicitly identified as rare in medical references.

\subsection{Setting II: Normal-to-Abnormal Discovery}
This setting aligns with the workflow of detecting pathology relative to \emph{diverse normal baselines}. We partition the label space into normal and abnormal subsets:
\begin{equation}
\mathcal{C}_{\mathrm{known}}=\mathcal{C}_{\mathrm{normal}}, \qquad
\mathcal{C}_{\mathrm{novel}}=\mathcal{C}_{\mathrm{abnormal}}, \qquad
\mathcal{C}_{\mathrm{normal}}\cap \mathcal{C}_{\mathrm{abnormal}}=\emptyset.
\end{equation}
The set $\mathcal{C}_{\mathrm{normal}}$ contains multiple normal subclasses (e.g., distinct anatomical landmarks), whereas $\mathcal{C}_{\mathrm{abnormal}}$ contains multiple pathological categories. By labeling $\mathcal{D}_l$ on $\mathcal{C}_{\mathrm{normal}}$, this setting evaluates a model's capacity to generalize beyond a \underline{normal manifold} to discover \emph{heterogeneous abnormal patterns}.

\subsection{Setting III: Within-Taxonomy Discovery}
This setting captures previously unseen subtypes \emph{within} established hierarchical taxonomies. Focusing on preterminal nodes, let $\mathcal{P}$ be the set of parent nodes whose children are exclusively leaf classes $\mathcal{L}$. We treat each sibling set as a semantic group: $\mathcal{S}(p)=\mathrm{children}(p)$ for $p\in\mathcal{P}$. We consider sibling groups with at least two leaf classes. For each $p\in\mathcal{P}$, let $n_p=|\mathcal{S}(p)|$ and $k_p=\min\{n_p-1,\max\{1,\lceil\rho n_p\rceil\}\}$. We choose $\mathcal{K}(p)\subseteq\mathcal{S}(p)$ with $|\mathcal{K}(p)|=k_p$, and define $\mathcal{N}(p)=\mathcal{S}(p)\setminus\mathcal{K}(p)$. This ensures at least one known and one novel class per group.
The global partitions are then defined as:
\begin{equation}
\mathcal{C}_{\mathrm{known}}=\bigcup_{p\in\mathcal{P}} \mathcal{K}(p), \qquad
\mathcal{C}_{\mathrm{novel}}=\bigcup_{p\in\mathcal{P}} \mathcal{N}(p).
\end{equation}
Smaller $\rho$ yields fewer labeled siblings per group and more novel siblings $\mathcal{N}(p)$ to discover. This mimics clinical taxonomies, testing the model's ability to isolate \underline{fine-grained novelty} that is \emph{semantically adjacent} to known diagnostic families.

\begin{figure}[tb]
  \centering
  \includegraphics[width=0.98\textwidth]{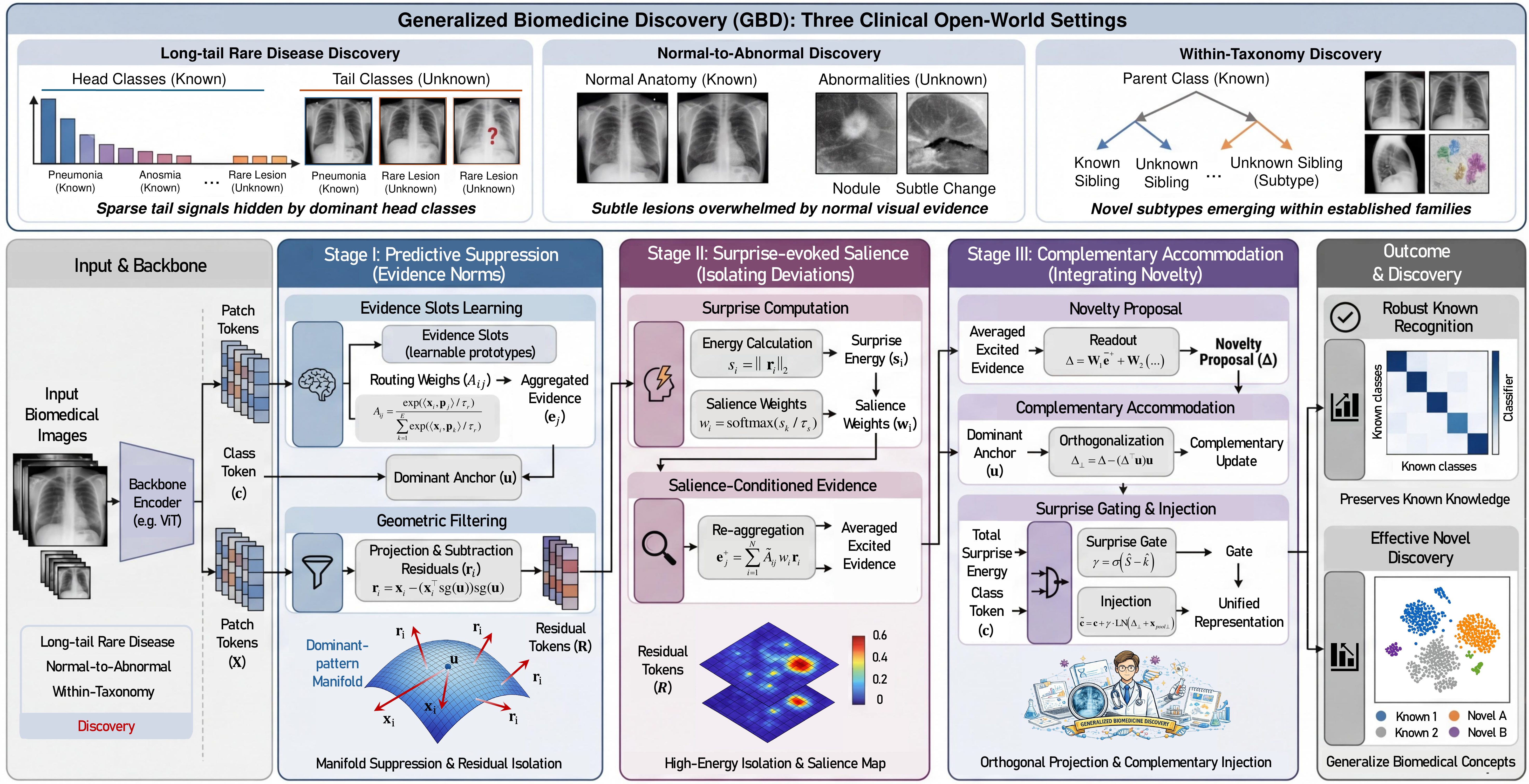}
\caption{\textbf{Method overview of SCAN} (Surprise-evoked Complementary AccommodatioN) for GBD. SCAN refines representations in three steps:
\textbf{(I)} it learns a notion of \emph{dominant evidence} and suppresses predictable visual patterns to expose residual signals;
\textbf{(II)} it scores the residuals by ``surprise'' to highlight sparse, high-energy deviations and aggregates them into salient evidence;
\textbf{(III)} it converts the salient evidence into a complementary update and injects it through a gated fusion, limiting disruption to known concepts while strengthening novel structure.}
  \label{fig:overview}
\end{figure}

\section{Method}
\label{sec:method}

Although the three GBD regimes differ clinically, they share a failure mode: recurring known structure dominates the global embedding, underrepresenting weak cues for rare classes, abnormalities, or new sibling subtypes. This motivates two requirements: expose deviations relative to each image's dominant content, then integrate them without indiscriminately rewriting established structure. We follow the analogous clinical progression of forming a predictive prior, attending to deviations, and accommodating informative evidence.
We instantiate this progression as \textbf{SCAN} in~\cref{fig:overview}, a lightweight plug-in that updates class and patch tokens through three stages:
Predictive Suppression~\cite{rao1999predictive}, Surprise-Evoked Salience~\cite{itti2009bayesian}, and Complementary Accommodation~\cite{mcclelland1995there}. Given an input image $\mathbf{x}\in\mathcal{X}$, an encoder outputs a class token and $N$ patch tokens:
\begin{equation}
\label{eq:scan_tokens}
\mathbf{c},\,\mathbf{X}=F_{\theta}(\mathbf{x}), \qquad
\mathbf{c}\in\mathbb{R}^{D},\ \mathbf{X}=[\mathbf{x}_1,\dots,\mathbf{x}_N]^{\top}\in\mathbb{R}^{N\times D}.
\end{equation}
SCAN performs a cognition-inspired transformation $(\mathbf{c},\mathbf{X})\mapsto \tilde{\mathbf{c}}$ in feature space, leaving the host discovery objective unchanged.

\subsection{Stage I: Predictive Suppression}
\label{sec:predictive_suppression}

The brain behaves like a ``predictor'': for \emph{expected} sensory content (e.g., dominant normal anatomy), feedback connections actively suppress predictable responses, freeing capacity for unexpected signals~\cite{rao1999predictive}.
In medical images, frequently occurring structures can form a strong manifold that suppresses subtle lesions and rare concepts; thus SCAN first builds a sample-conditioned \emph{dominant anchor} and geometrically filters out predictable components.

\paragraph{Dominant evidence and dominant anchor.}
We softly summarize recurring visual patterns by routing patch tokens into $E$ evidence slots, which serve as a compact, image-adaptive dominant-pattern codebook.
Let $\mathbf{P}\in\mathbb{R}^{E\times D}$ be learnable evidence prototypes and $\tau_r$ a temperature. We compute routing weights:
\begin{equation}
\label{eq:routing_evidence}
A_{ij}=\frac{\exp(\langle \mathbf{x}_i,\mathbf{p}_j\rangle/\tau_r)}{\sum_{k=1}^{E}\exp(\langle \mathbf{x}_i,\mathbf{p}_k\rangle/\tau_r)}.
\end{equation}
We renormalize over tokens so that each slot forms a weighted visual summary independent of its assigned token mass:
\begin{equation}
\label{eq:slot_renorm}
\tilde{A}_{ij}=\frac{A_{ij}}{\sum_{k=1}^{N}A_{kj}},\qquad
\mathbf{e}_j=\sum_{i=1}^{N}\tilde{A}_{ij}\,\mathbf{x}_i,\qquad
\bar{\mathbf{e}}=\frac{1}{E}\sum_{j=1}^{E}\mathbf{e}_j,
\end{equation}
where $\langle\cdot,\cdot\rangle$ denotes dot product (cosine if tokens are normalized).
We then form a sample-specific \emph{dominant anchor} direction:
\begin{equation}
\label{eq:normal_anchor}
\mathbf{u}=\frac{\mathbf{c}+\bar{\mathbf{e}}}{\|\mathbf{c}+\bar{\mathbf{e}}\|_2}.
\end{equation}
Fusing global and mean-slot evidence yields an image-adaptive anchor instead of a fixed normal prototype.

\paragraph{Predictive suppression by geometric filtering.}
We actively remove the predictable component of each patch token along the dominant anchor, producing residuals in the orthogonal complement:
\begin{equation}
\label{eq:predictive_suppression}
\mathbf{r}_i=\mathbf{x}_i-(\mathbf{x}_i^{\top}\,\mathrm{sg}(\mathbf{u}))\,\mathrm{sg}(\mathbf{u}),\qquad
\mathbf{R}=[\mathbf{r}_1,\dots,\mathbf{r}_N]^{\top}.
\end{equation}
Here $\mathrm{sg}(\cdot)$ denotes stop-gradient and stabilizes the anchor. The projection removes only its aligned component and preserves the remaining token geometry, requiring neither normal labels nor anomaly templates.

\subsection{Stage II: Surprise-Evoked Salience}
\label{sec:surprise_salience}

After suppression, the residual norm measures image-conditional mismatch: a token is surprising when the dominant anchor explains it poorly. This mirrors the attentional role of Bayesian surprise~\cite{itti2009bayesian} without requiring a dataset-specific anomaly threshold.

\paragraph{Surprise energy and salience weights.}
We quantify patch-wise surprise by the residual energy and convert it into a salience distribution:
\begin{equation}
\label{eq:surprise_salience_weights}
s_i=\|\mathbf{r}_i\|_2,\qquad
w_i=\frac{\exp(s_i/\tau_s)}{\sum_{k=1}^{N}\exp(s_k/\tau_s)},
\end{equation}
where $\tau_s$ controls concentration. The within-image softmax ranks residual evidence without assuming cross-dataset calibration.

\paragraph{Salience-conditioned evidence of novelty.}
We re-aggregate residuals using both their original slot assignments and their surprise weights, yielding \emph{excited} evidence that concentrates on coherent, unpredicted deviations:
\begin{equation}
\label{eq:excited_evidence}
\mathbf{e}^{+}_j=\sum_{i=1}^{N}\tilde{A}_{ij}\,w_i\,\mathbf{r}_i,\qquad
\bar{\mathbf{e}}^{+}=\frac{1}{E}\sum_{j=1}^{E}\mathbf{e}^{+}_j.
\end{equation}
Here, $\tilde{A}_{ij}$ preserves slot semantics, whereas $w_i$ controls residual contribution. Eqs.~(\ref{eq:surprise_salience_weights})--(\ref{eq:excited_evidence}) thus separate \emph{what recurring pattern a token belongs to} from \emph{how unexpectedly it departs}.

\paragraph{Token refinement (broadcast of excited evidence).}
Beyond global evidence, clinicians also refine a local ``mental map'' of the image.
We broadcast excited evidence back to tokens with a lightweight readout:
\begin{equation}
\label{eq:token_refine}
\mathbf{X}^{out}=\mathbf{X}+\phi(\mathbf{A}\mathbf{E}^{+}),\qquad
\mathbf{E}^{+}=[\mathbf{e}^{+}_1,\dots,\mathbf{e}^{+}_E]^{\top}\in\mathbb{R}^{E\times D},
\end{equation}
where $\phi(\cdot)$ is a small MLP applied row-wise.

\subsection{Stage III: Complementary Accommodation}
\label{sec:complementary_accommodation}

Discovery must integrate deviations into the global representation, yet an unconstrained update may distort known structure. Motivated by Complementary Learning Systems theory~\cite{mcclelland1995there}, SCAN constructs an additive update with reduced overlap with the dominant direction.

\paragraph{Novelty vector from excited evidence.}
We map excited evidence into a global novelty proposal via a simple first- and second-order readout:
\begin{equation}
\label{eq:novelty_readout}
\Delta=\mathbf{W}_1\,\bar{\mathbf{e}}^{+}+\mathbf{W}_2\left(\frac{1}{E}\sum_{j=1}^{E}\mathbf{e}^{+}_j\odot \mathbf{e}^{+}_j\right),
\end{equation}
where $\odot$ is the Hadamard product and $\mathbf{W}_1,\mathbf{W}_2\in\mathbb{R}^{D\times D}$ are learnable.
The first-order term retains the signed mean shift; the second records dimension-wise energy that may cancel in the mean. Together, they capture consistent and sparse-but-strong deviations without another attention block.

\paragraph{Complementarity.}
We reduce direct interference with the dominant content by removing the component of the proposal aligned with the anchor:
\begin{equation}
\label{eq:complementary_projection}
\Delta_{\perp}=\Delta-(\Delta^{\top}\mathbf{u})\mathbf{u}.
\end{equation}
We further summarize refined tokens by $\mathbf{x}_{pool}=\mathrm{MeanPool}(\mathbf{X}^{out})\in\mathbb{R}^{D}$ and project this token evidence into the same complementary subspace:
\begin{equation}
\label{eq:token_perp}
\mathbf{x}_{pool\perp}=\mathbf{x}_{pool}-(\mathbf{x}_{pool}^{\top}\mathbf{u})\mathbf{u}.
\end{equation}
Projecting both branches into the same subspace prevents the pooled token summary from reintroducing the dominant component removed from the global proposal.

\paragraph{Non-parametric surprise gate.}
Rather than learning dataset-specific scalar calibrations, we define $\gamma$ from three cognitively motivated signals:
(i) \emph{surprise magnitude} $S=\sum_i w_i s_i$,
(ii) \emph{knownness} $k=\cos(\mathbf{c},\bar{\mathbf{e}})$ (how well the sample aligns with dominant evidence),
and (iii) \emph{salience concentration} via peakedness $\kappa$:
\begin{equation}
\label{eq:gate_terms}
S=\sum_{i=1}^{N}w_i s_i,\qquad
k=\cos(\mathbf{c},\bar{\mathbf{e}}),\qquad
\kappa = 1-\frac{\mathrm{Ent}(\mathbf{w})}{\log N},\ \ \mathrm{Ent}(\mathbf{w})=-\sum_{i=1}^{N}w_i\log w_i.
\end{equation}
Using EMA moments $(\mu_S,\sigma_S)$ and $(\mu_k,\sigma_k)$, we standardize the signals as $\hat{S}=(S-\mu_S)/\sigma_S$ and $\hat{k}=(k-\mu_k)/\sigma_k$. The robust gate is $\gamma=\sigma(\hat{S}-\hat{k})\cdot\mathrm{clip}(\kappa,0,1)$. The three factors are complementary: $S$ measures deviation strength, $k$ discounts samples already explained by dominant evidence, and $\kappa$ suppresses diffuse residual activation. Accommodation is therefore selective in both residual magnitude and salience concentration.

\paragraph{Stable accommodation.}
We assimilate novelty as a complementary residual:
\begin{equation}
\label{eq:final_update}
\tilde{\mathbf{c}}=\mathbf{c}+\gamma \cdot \mathrm{LN}\!\left(\Delta_{\perp}+\mathbf{x}_{pool\perp}\right),
\end{equation}
The identity path recovers the original representation when $\gamma$ is small, and $\mathrm{LN}(\cdot)$ controls update scale. This cannot guarantee complete independence from known features, but it biases accommodation away from the dominant direction and limits feature drift. The updated $\tilde{\mathbf{c}}$ enters the unchanged GCD pipeline.

\paragraph{Connection to the GBD.}
The anchor has a setting-specific interpretation but a shared role. In Setting I, it summarizes head-class patterns, exposing underrepresented tail cues. In Setting II, it reflects recurring normal anatomy, retaining localized pathology in the residual. In Setting III, it captures parent-level morphology, while accommodation emphasizes subtype differences. SCAN thereby targets the common suppression mechanism without setting-specific supervision.

\section{Experiments}
\label{sec:mainexp}

Through comprehensive experiments, we seek to answer three core questions:
(1) Benchmark validity. Does GBD capture the key open-world challenges in medical imaging, and can it differentiate methods across long-tail, anomaly-dominated, and taxonomy-aware settings?
(2) Effectiveness and generality. Does SCAN improve generalized biomedicine discovery, and can it serve as a lightweight plug-and-play module that strengthens diverse host baselines without altering their original training protocols?
(3) Mechanism and practicality. Which components drive the gains, how do they interact in the full framework, and how robust is SCAN under key hyperparameter choices and compute budgets?



\subsection{Experimental Setup}

\paragraph{\textbf{Datasets.}}
We evaluate GBD on four publicly available datasets spanning diverse biomedical imaging domains.
Each dataset provides an explicit class label per image (either by folder structure or an official label file), and several of them additionally provide a native taxonomy used for our GBD's settings. \textbf{Gastrovision}~\cite{jha2023gastrovision} (endoscopy): multi-center GI endoscopy dataset spanning upper and lower GI tracts; labels are organized by clinically meaningful finding types, including anatomical landmarks/normal findings, pathological abnormalities, and procedure-related categories.
\textbf{HistoSet-5$\times$14}~\cite{rafi2025histoset} (histopathology): multi-organ H\&E collection covering five organs (breast, colon, lung, oral cavity, ovary); the label space is naturally structured by organ identity and tissue state across 14 tissue classes.
\textbf{MLL23}~\cite{mll23} (microscopy): expert-annotated peripheral blood single-cell images; labels follow hematopoietic structure with major groups and finer subdivisions into mature and immature cell types/stages.
\textbf{DERM12345}~\cite{yilmaz2024derm12345} (dermatology): dermatoscopic skin lesion dataset with a taxonomic tree comprising 5 super-classes, 15 main classes, and 40 subclasses.

\begin{table}[t]
    \caption{Comparison on \textbf{Setting I} (Long-tail Rare Disease Discovery) across \colorbox{yellow!5}{\color{black}endoscopy}, \colorbox{red!5}{\color{black}pathology}, \colorbox{blue!5}{\color{black}microscopy}, and \colorbox{green!5}{\color{black}dermatology} images. Results are accuracy on All, Old, and New subsets, with the best results highlighted in bold.}
    \vspace{-0.5cm}
  \label{tab:setting1}
  \begin{center}
     \scalebox{0.79}{
 \renewcommand\tabcolsep{3pt}
    \begin{tabular}[b]{p{0.15\textwidth} ccc ccc ccc ccc ccc}
    \toprule
        \multirow{2}{*}{Method} &
        \multicolumn{3}{c}{\cellcolor{yellow!5}\textbf{Gastrovision}} &
        \multicolumn{3}{c}{\cellcolor{red!5}\textbf{HistoSet}} &
        \multicolumn{3}{c}{\cellcolor{blue!5}\textbf{MLL23}} &
        \multicolumn{3}{c}{\cellcolor{green!5}\textbf{Derm12345}} &
        \multicolumn{3}{c}{\textbf{Average}} \\
        \cmidrule(lr){2-4} \cmidrule(lr){5-7} \cmidrule(lr){8-10}
        \cmidrule(lr){11-13} \cmidrule(lr){14-16}
        & All & Old & New & All & Old & New & All & Old & New
        & All & Old & New & All & Old & New \\

    \midrule

        SEAL$^{\dag}$ & 56.8 & 51.2 & \textbf{60.7} & \textbf{81.2} & 87.1 & 77.9 & 65.4 & 67.6 & 54.1 & \textbf{56.5} & \textbf{61.0} & 10.6 & 65.0 & 66.7 & 50.8 \\
        \midrule

        SelEx & 50.1 & 52.0 & 47.4 & 76.4 & 97.5 & 70.5 & 66.4 & 68.3 & 61.3 & 28.9 & 29.1 & 27.9 & 55.5 & 61.7 & 51.8\\
         \rowcolor{gray!15}\quad+ SCAN & \textbf{71.2} & \textbf{83.4} & 53.0 & 76.3 & \textbf{99.9} & 69.7 & \textbf{80.7} & \textbf{89.3} & 58.6 & 35.6 & 35.6 & \textbf{35.6} & \textbf{66.0} & \textbf{77.1} & \textbf{54.2}\\
          \quad $\Delta$ & \textbf{\textcolor{ForestGreen}{21.1}} & \textbf{\textcolor{ForestGreen}{31.4}} & \textbf{\textcolor{ForestGreen}{5.6}} & \textcolor{red!50!white}{-0.1} & \textbf{\textcolor{ForestGreen}{2.4}} & \textcolor{red!50!white}{-0.8} & \textbf{\textcolor{ForestGreen}{14.3}} & \textbf{\textcolor{ForestGreen}{20.9}} & \textcolor{red!50!white}{-2.7} & \textbf{\textcolor{ForestGreen}{6.7}} & \textbf{\textcolor{ForestGreen}{6.5}} & \textbf{\textcolor{ForestGreen}{7.7}} & \textbf{\textcolor{ForestGreen}{10.5}} & \textbf{\textcolor{ForestGreen}{15.3}} & \textbf{\textcolor{ForestGreen}{2.4}} \\
          
        SimGCD & 37.8 & 40.6 & 31.8 & 74.4 & 83.8 & 71.8 & 56.9 & 52.9 & 67.0 & 25.4 & 27.9 & 13.1 & 48.6 & 51.3 & 45.9 \\
         \rowcolor{gray!15} \quad+ SCAN & 39.7 & 43.3 & 31.9 & 80.0 & 82.1 & \textbf{79.4} & 64.3 & 63.1 & 67.2  & 25.9 & 28.2 & 14.2 & 52.5 & 54.2 & 48.1\\
          \quad $\Delta$ & \textbf{\textcolor{ForestGreen}{1.9}} & \textbf{\textcolor{ForestGreen}{2.7}} & \textbf{\textcolor{ForestGreen}{0.1}} & \textbf{\textcolor{ForestGreen}{5.6}} & \textcolor{red!50!white}{-1.7} & \textbf{\textcolor{ForestGreen}{7.6}} & \textbf{\textcolor{ForestGreen}{7.4}} & \textbf{\textcolor{ForestGreen}{10.2}} & \textbf{\textcolor{ForestGreen}{0.2}} & \textbf{\textcolor{ForestGreen}{0.5}} & \textbf{\textcolor{ForestGreen}{0.4}} & \textbf{\textcolor{ForestGreen}{1.1}} & \textbf{\textcolor{ForestGreen}{3.8}} & \textbf{\textcolor{ForestGreen}{2.9}} & \textbf{\textcolor{ForestGreen}{2.2}} \\
          
        LegoGCD & 39.8 & 48.9 & 26.1 & 72.5 & 76.0 & 71.6 & 70.8 & 63.4 & \textbf{89.7} & 26.3 & 28.5 & 15.7 & 52.3 & 54.2 & 50.8 \\
         \rowcolor{gray!15} \quad+ SCAN & 41.3 & 49.4 & 29.2 & 72.5 & 77.5 & 71.2 & 73.2 & 67.2 & 88.8  & 26.7 & 29.0 & 14.8 & 53.4 & 55.8 & 51.0\\
          \quad $\Delta$ & \textbf{\textcolor{ForestGreen}{1.5}} & \textbf{\textcolor{ForestGreen}{0.5}} & \textbf{\textcolor{ForestGreen}{3.1}} & \textbf{\textcolor{ForestGreen}{0.0}} & \textbf{\textcolor{ForestGreen}{1.5}} & \textcolor{red!50!white}{-0.4} & \textbf{\textcolor{ForestGreen}{2.4}} & \textbf{\textcolor{ForestGreen}{3.8}} & \textcolor{red!50!white}{-0.9} & \textbf{\textcolor{ForestGreen}{0.3}} & \textbf{\textcolor{ForestGreen}{0.6}} & \textcolor{red!50!white}{-0.9} & \textbf{\textcolor{ForestGreen}{1.1}} & \textbf{\textcolor{ForestGreen}{1.6}} & \textbf{\textcolor{ForestGreen}{0.2}} \\

\midrule
           \quad \textbf{Avg.} $\Delta$ & \textbf{\textcolor{ForestGreen}{8.2}} & \textbf{\textcolor{ForestGreen}{11.5}} & \textbf{\textcolor{ForestGreen}{2.9}} & \textbf{\textcolor{ForestGreen}{1.8}} & \textbf{\textcolor{ForestGreen}{0.7}} & \textbf{\textcolor{ForestGreen}{2.1}} & \textbf{\textcolor{ForestGreen}{8.1}} & \textbf{\textcolor{ForestGreen}{11.6}} & \textcolor{red!50!white}{-1.2} & \textbf{\textcolor{ForestGreen}{2.5}} & \textbf{\textcolor{ForestGreen}{2.5}} & \textbf{\textcolor{ForestGreen}{2.7}} & \textbf{\textcolor{ForestGreen}{5.1}} & \textbf{\textcolor{ForestGreen}{6.6}} & \textbf{\textcolor{ForestGreen}{1.6}} \\
    \bottomrule
    \end{tabular}}
\end{center}
\end{table}

\begin{table}[t]
    \caption{Comparison on \textbf{Setting II} (Normal-to-Abnormal Discovery) across \colorbox{yellow!5}{\color{black}endoscopy}, \colorbox{red!5}{\color{black}pathology}, \colorbox{blue!5}{\color{black}microscopy}, and \colorbox{green!5}{\color{black}dermatology} images. Results are accuracy on All, Old, and New subsets, with the best results highlighted in bold.}
    \vspace{-0.5cm}
  \label{tab:setting2}
  \begin{center}
     \scalebox{0.79}{
 \renewcommand\tabcolsep{3pt}
    \begin{tabular}[b]{p{0.15\textwidth} ccc ccc ccc ccc ccc}
    \toprule
        \multirow{2}{*}{Method} &
        \multicolumn{3}{c}{\cellcolor{yellow!5}\textbf{Gastrovision}} &
        \multicolumn{3}{c}{\cellcolor{red!5}\textbf{HistoSet}} &
        \multicolumn{3}{c}{\cellcolor{blue!5}\textbf{MLL23}} &
        \multicolumn{3}{c}{\cellcolor{green!5}\textbf{Derm12345}} &
        \multicolumn{3}{c}{\textbf{Average}} \\
        \cmidrule(lr){2-4} \cmidrule(lr){5-7} \cmidrule(lr){8-10}
        \cmidrule(lr){11-13} \cmidrule(lr){14-16}
        & All & Old & New & All & Old & New & All & Old & New
        & All & Old & New & All & Old & New \\

    \midrule

        SEAL$^{\dag}$ & 62.6 & 64.2 & \textbf{61.0} & \textbf{73.5} & \textbf{98.7} & 48.2 & 69.2 & 70.9 & 60.1 & \textbf{59.0} & \textbf{61.7} & 5.5 & \textbf{66.1} & 73.8 & 43.7 \\
        \midrule

        SelEx & 50.3 & 51.9 & 41.7 & 64.3 & 97.7 & 47.5 & 58.5 & 58.0 & 59.8 & 29.3 & 28.2 & \textbf{40.7} & 50.6 & 58.9 & 47.4\\
         \rowcolor{gray!15}\quad+ SCAN & \textbf{70.1} & \textbf{86.0} & 46.3 & 68.6 & 97.2 & \textbf{60.7} & \textbf{87.1} & \textbf{89.1} & \textbf{81.7} & 33.6 & 33.3 & 36.4 & 64.8 & \textbf{76.4} & \textbf{56.3}\\
          \quad $\Delta$ & \textbf{\textcolor{ForestGreen}{19.8}} & \textbf{\textcolor{ForestGreen}{34.1}} & \textbf{\textcolor{ForestGreen}{4.6}} & \textbf{\textcolor{ForestGreen}{4.4}} & \textcolor{red!50!white}{-0.5} & \textbf{\textcolor{ForestGreen}{13.1}} & \textbf{\textcolor{ForestGreen}{28.6}} & \textbf{\textcolor{ForestGreen}{31.1}} & \textbf{\textcolor{ForestGreen}{21.9}} & \textbf{\textcolor{ForestGreen}{4.3}} & \textbf{\textcolor{ForestGreen}{5.1}} & \textcolor{red!50!white}{-4.3} & \textbf{\textcolor{ForestGreen}{14.2}} & \textbf{\textcolor{ForestGreen}{17.5}} & \textbf{\textcolor{ForestGreen}{8.8}} \\

        SimGCD & 39.7 & 42.9 & 25.6 & 61.1 & 94.1 & 44.5 & 57.9 & 55.9 & 63.4 & 26.6 & 27.2 & 19.9 & 46.3 & 55.0 & 38.3 \\
         \rowcolor{gray!15} \quad+ SCAN & 40.1 & 44.0 & 22.6 & 62.0 & 93.6 & 46.1 & 58.3 & 57.5 & 60.6 & 26.4 & 26.8 & 21.6 & 46.7 & 55.5 & 37.7\\
          \quad $\Delta$ & \textbf{\textcolor{ForestGreen}{0.4}} & \textbf{\textcolor{ForestGreen}{1.1}} & \textcolor{red!50!white}{-3.0} & \textbf{\textcolor{ForestGreen}{0.9}} & \textcolor{red!50!white}{-0.6} & \textbf{\textcolor{ForestGreen}{1.6}} & \textbf{\textcolor{ForestGreen}{0.5}} & \textbf{\textcolor{ForestGreen}{1.7}} & \textcolor{red!50!white}{-2.8} & \textcolor{red!50!white}{-0.2} & \textcolor{red!50!white}{-0.4} & \textbf{\textcolor{ForestGreen}{1.7}} & \textbf{\textcolor{ForestGreen}{0.4}} & \textbf{\textcolor{ForestGreen}{0.5}} & \textcolor{red!50!white}{-0.6} \\

        LegoGCD & 49.3 & 53.7 & 24.9 & 62.0 & 97.5 & 44.2 & 68.5 & 72.1 & 58.6 & 28.5 & 30.8 & 5.5 & 52.1 & 63.5 & 33.3 \\
         \rowcolor{gray!15} \quad+ SCAN & 50.5 & 55.5 & 23.1 & 63.9 & 97.4 & 47.1 & 69.4 & 73.7 & 57.7 & 29.0 & 29.7 & 21.6 & 53.2 & 64.1 & 37.4\\
          \quad $\Delta$ & \textbf{\textcolor{ForestGreen}{1.3}} & \textbf{\textcolor{ForestGreen}{1.8}} & \textcolor{red!50!white}{-1.8} & \textbf{\textcolor{ForestGreen}{1.9}} & \textcolor{red!50!white}{-0.1} & \textbf{\textcolor{ForestGreen}{2.9}} & \textbf{\textcolor{ForestGreen}{0.9}} & \textbf{\textcolor{ForestGreen}{1.6}} & \textcolor{red!50!white}{-0.9} & \textbf{\textcolor{ForestGreen}{0.4}} & \textcolor{red!50!white}{-1.1} & \textbf{\textcolor{ForestGreen}{16.1}} & \textbf{\textcolor{ForestGreen}{1.1}} & \textbf{\textcolor{ForestGreen}{0.6}} & \textbf{\textcolor{ForestGreen}{4.1}} \\

\midrule
           \quad \textbf{Avg.} $\Delta$ & \textbf{\textcolor{ForestGreen}{7.1}} & \textbf{\textcolor{ForestGreen}{12.3}} & \textcolor{red!50!white}{-0.1} & \textbf{\textcolor{ForestGreen}{2.4}} & \textcolor{red!50!white}{-0.4} & \textbf{\textcolor{ForestGreen}{5.9}} & \textbf{\textcolor{ForestGreen}{10.0}} & \textbf{\textcolor{ForestGreen}{11.5}} & \textbf{\textcolor{ForestGreen}{6.1}} & \textbf{\textcolor{ForestGreen}{1.5}} & \textbf{\textcolor{ForestGreen}{1.2}} & \textbf{\textcolor{ForestGreen}{4.5}} & \textbf{\textcolor{ForestGreen}{5.2}} & \textbf{\textcolor{ForestGreen}{6.2}} & \textbf{\textcolor{ForestGreen}{4.0}} \\
    \bottomrule
    \end{tabular}}
\end{center}
\end{table}

\begin{table}[t]
    \caption{Comparison on \textbf{Setting III} (Within-Taxonomy Discovery) across \colorbox{yellow!5}{\color{black}endoscopy}, \colorbox{red!5}{\color{black}pathology}, \colorbox{blue!5}{\color{black}microscopy}, and \colorbox{green!5}{\color{black}dermatology} images. Results are accuracy on All, Old, and New subsets, with the best results highlighted in bold.}
        \vspace{-0.5cm}
  \label{tab:setting3}
  \begin{center}
     \scalebox{0.79}{
 \renewcommand\tabcolsep{3pt}
    \begin{tabular}[b]{p{0.15\textwidth} ccc ccc ccc ccc ccc}
    \toprule
        \multirow{2}{*}{Method} &
        \multicolumn{3}{c}{\cellcolor{yellow!5}\textbf{Gastrovision}} &
        \multicolumn{3}{c}{\cellcolor{red!5}\textbf{HistoSet}} &
        \multicolumn{3}{c}{\cellcolor{blue!5}\textbf{MLL23}} &
        \multicolumn{3}{c}{\cellcolor{green!5}\textbf{Derm12345}} &
        \multicolumn{3}{c}{\textbf{Average}} \\
        \cmidrule(lr){2-4} \cmidrule(lr){5-7} \cmidrule(lr){8-10}
        \cmidrule(lr){11-13} \cmidrule(lr){14-16}
        & All & Old & New & All & Old & New & All & Old & New
        & All & Old & New & All & Old & New \\

    \midrule

        SEAL$^{\dag}$ & 54.9 & 51.3 & \textbf{64.3} & \textbf{92.9} & 95.7 & \textbf{89.3} & 70.6 & 83.3 & 47.2 & 27.7 & 62.4 & 16.4 & 61.5 & 73.2 & \textbf{54.3} \\
        \midrule

        SelEx & 50.1 & 45.7 & 53.8 & 77.5 & \textbf{99.4} & 71.4 & 62.5 & 69.9 & 55.6 & 24.8 & 65.1 & 18.2 & 53.7 & 70.0 & 49.8 \\
         \rowcolor{gray!15}\quad+ SCAN & \textbf{66.2} & \textbf{79.6} & 46.3 & 88.5 & 96.4 & 83.2 & \textbf{81.4} & \textbf{89.0} & \textbf{61.9} & \textbf{27.8} & \textbf{69.9} & \textbf{20.9} & \textbf{66.0} & \textbf{83.7} & 53.1\\
          \quad $\Delta$ & \textbf{\textcolor{ForestGreen}{16.1}} & \textbf{\textcolor{ForestGreen}{33.8}} & \textcolor{red!50!white}{-7.5} & \textbf{\textcolor{ForestGreen}{11.0}} & \textcolor{red!50!white}{-3.0} & \textbf{\textcolor{ForestGreen}{11.7}} & \textbf{\textcolor{ForestGreen}{18.9}} & \textbf{\textcolor{ForestGreen}{19.0}} & \textbf{\textcolor{ForestGreen}{6.3}} & \textbf{\textcolor{ForestGreen}{3.0}} & \textbf{\textcolor{ForestGreen}{4.8}} & \textbf{\textcolor{ForestGreen}{2.7}} & \textbf{\textcolor{ForestGreen}{12.3}} & \textbf{\textcolor{ForestGreen}{13.7}} & \textbf{\textcolor{ForestGreen}{3.3}} \\

        SimGCD & 42.9 & 43.0 & 42.5 & 89.4 & 91.7 & 87.9 & 53.2 & 57.4 & 49.4 & 19.3 & 40.8 & 15.9 & 51.2 & 58.2 & 48.9 \\
         \rowcolor{gray!15} \quad+ SCAN & 42.0 & 42.4 & 39.4 & 89.1 & 90.5 & 88.2 & 54.0 & 58.4 & 49.9  & 17.8 & 36.8 & 14.7 & 50.7 & 57.0 & 48.1\\
          \quad $\Delta$ & \textcolor{red!50!white}{-0.9} & \textcolor{red!50!white}{-0.6} & \textcolor{red!50!white}{-3.1} & \textcolor{red!50!white}{-0.3} & \textcolor{red!50!white}{-1.2} & \textbf{\textcolor{ForestGreen}{0.2}} & \textbf{\textcolor{ForestGreen}{0.8}} & \textbf{\textcolor{ForestGreen}{1.1}} & \textbf{\textcolor{ForestGreen}{0.5}} & \textcolor{red!50!white}{-1.5} & \textcolor{red!50!white}{-4.0} & \textcolor{red!50!white}{-1.2} & \textcolor{red!50!white}{-0.5} & \textcolor{red!50!white}{-1.2} & \textcolor{red!50!white}{-0.8} \\

        LegoGCD & 40.2 & 39.6 & 40.6 & 75.8 & 84.4 & 70.0 & 54.7 & 61.8 & 48.2 & 19.4 & 41.0 & 16.0 & 47.5 & 56.7 & 43.7 \\
         \rowcolor{gray!15} \quad+ SCAN & 40.5 & 40.4 & 40.6 & 80.5 & 86.2 & 76.6 & 60.3 & 68.7 & 52.6 & 20.8 & 42.4 & 17.2 & 50.5 & 59.4 & 46.8\\
          \quad $\Delta$ & \textbf{\textcolor{ForestGreen}{0.4}} & \textbf{\textcolor{ForestGreen}{0.8}} & \textbf{\textcolor{ForestGreen}{0.0}} & \textbf{\textcolor{ForestGreen}{4.7}} & \textbf{\textcolor{ForestGreen}{1.8}} & \textbf{\textcolor{ForestGreen}{6.6}} & \textbf{\textcolor{ForestGreen}{5.6}} & \textbf{\textcolor{ForestGreen}{6.9}} & \textbf{\textcolor{ForestGreen}{4.4}} & \textbf{\textcolor{ForestGreen}{1.3}} & \textbf{\textcolor{ForestGreen}{1.5}} & \textbf{\textcolor{ForestGreen}{1.3}} & \textbf{\textcolor{ForestGreen}{3.0}} & \textbf{\textcolor{ForestGreen}{2.8}} & \textbf{\textcolor{ForestGreen}{3.1}} \\

\midrule
           \quad \textbf{Avg.} $\Delta$ & \textbf{\textcolor{ForestGreen}{5.2}} & \textbf{\textcolor{ForestGreen}{11.3}} & \textcolor{red!50!white}{-3.6} & \textbf{\textcolor{ForestGreen}{5.1}} & \textcolor{red!50!white}{-0.8} & \textbf{\textcolor{ForestGreen}{6.2}} & \textbf{\textcolor{ForestGreen}{8.4}} & \textbf{\textcolor{ForestGreen}{9.0}} & \textbf{\textcolor{ForestGreen}{3.7}} & \textbf{\textcolor{ForestGreen}{0.9}} & \textbf{\textcolor{ForestGreen}{0.8}} & \textbf{\textcolor{ForestGreen}{0.9}} & \textbf{\textcolor{ForestGreen}{4.9}} & \textbf{\textcolor{ForestGreen}{5.1}} & \textbf{\textcolor{ForestGreen}{1.8}} \\
    \bottomrule
    \end{tabular}}
\end{center}
\end{table}


\paragraph{\textbf{Evaluation protocol.}}
For each dataset, we train models on $\mathcal{D}_l\cup\mathcal{D}_u$ \emph{without access to the ground-truth labels $y_i^u$ in $\mathcal{D}_u$}.
At test time, we measure clustering accuracy between the $y_i^u$ and the model predictions $\hat{y}_i$ over $\mathcal{D}_u$ as:
\begin{equation}
\label{eq:acc}
\mathrm{ACC} = \max_{p\in \mathcal{P}(\mathcal{Y}_u)} \frac{1}{|\mathcal{D}_u|}\sum_{i=1}^{|\mathcal{D}_u|}\mathbb{1}\left\{y_i^u = p(\hat{y}_i)\right\}.
\end{equation}
where $\mathcal{P}(\mathcal{Y}_u)$ is the set of all permutations of class labels in the unlabeled label space $\mathcal{Y}_u$.
The maximization is computed via the Hungarian optimal assignment algorithm.
Following standard practice in generalized discovery, we report $\mathrm{ACC}$ on all unlabeled instances, and additionally on the known subset (instances in $\mathcal{D}_u$ with labels in $\mathcal{C}_{\mathrm{known}}$) and the novel subset (instances in $\mathcal{D}_u$ with labels in $\mathcal{C}_{\mathrm{novel}}$).
Importantly, we compute the Hungarian assignment \emph{once} over all classes in $\mathcal{Y}_u$, and only then evaluate the known/novel subset accuracies under the same assignment.
Following the standard generalized discovery protocol~\cite{gcd}, we assume the number of novel classes $K_{\mathrm{novel}}$ is available for evaluation.


\paragraph{\textbf{Implementation details.}}
We implement SCAN as plug-and-play that can be seamlessly attached to diverse GCD baselines. All methods use a ViT-B/16 backbone initialized with self-supervised DINO-V2 weights~\cite{oquab2023dinov2}. Models are trained for $200$ epochs. In our default configuration, SCAN uses $E=16$ evidence slots.
To evaluate the generality of our approach, we largely follow the \textbf{\emph{original training hyperparameters and optimization schedules}} of each host baseline, without method-specific tuning. Following the original GCD configuration~\cite{gcd}, we allocate 50\% of the samples from the known set to the unlabeled subset. We generally use $\alpha=0.5$ in GBD's Setting I and use $\rho=0.5$ in Setting III.

\paragraph{\textbf{Baselines.}}
We test SCAN with representative GCD baselines for forgetting mitigation and hierarchy-aware learning.
  \begin{itemize}
    \item \textbf{SimGCD}~\cite{simgcd}: a simple one-stage parametric classifier for GCD, using debiased learning and entropy regularization to reduce the tendency.
    \item \textbf{LegoGCD}~\cite{legogcd}: a modular add-on that targets catastrophic forgetting in GCD by regularizing predictions to better preserve known-class knowledge while improving novel discrimination.
    \item \textbf{SelEx}~\cite{selex}: a fine-grained GCD method that builds ``self-expertise'' via hierarchical pseudo-labeling and complementary supervised/unsupervised training signals to separate subtle categories.
    \item \textbf{SEAL}~\cite{he2025seal}: a semantic-aware hierarchical framework that leverages naturally available hierarchies. Because SEAL requires additional hierarchical labels, we treat it as an auxiliary performance reference rather than a directly comparable baseline.
  \end{itemize}

\subsection{Main Results}
\label{subsec:exp_Main_Results}

\paragraph{\textbf{Results on Long-tail Rare Disease Discovery.}}
\Cref{tab:setting1} evaluates SCAN on the clinically critical long-tail regime across four imaging domains, where class-frequency imbalance leaves tail categories sparsely represented relative to dominant head classes. SCAN improves the average \textit{All} accuracy of all three host baselines, yielding a notable \textbf{+10.5\%} gain on SelEx, although several subset-specific regressions remain. The effect is most pronounced on Gastrovision, where \textit{Old} accuracy rises from 52.0\% to \textbf{83.4\%} (\textbf{+31.4\%}), while on Derm12345 it boosts tail recognition from 27.9\% to 35.6\% (\textbf{+7.7\%}). Although SEAL is not directly comparable because it uses additional hierarchical labels, SelEx+SCAN exceeds its reported accuracy in several settings, suggesting that SCAN can recover part of the benefit of hierarchy-aware supervision without explicitly using such labels. These results support our motivation: \emph{predictive suppression} attenuates dominant head-class patterns and \emph{surprise-evoked salience} amplifies underrepresented tail cues.

\paragraph{\textbf{Results on Normal-to-Abnormal Discovery.}}
\Cref{tab:setting2} shows that SCAN generally strengthens normal-to-abnormal discovery across the four imaging domains, a clinically crucial setting where normal anatomy can suppress faint pathological cues. With the strong SelEx baseline, SCAN improves the average accuracy by +14.2\% on \textit{All} and +8.8\% on abnormal \textit{New}, and the impact is most pronounced on Gastrovision, where \textit{Old} increases by \textbf{34.1\%}, and on MLL23, where \textit{All} and \textit{New} rise by \textbf{28.6\%} and \textbf{21.9\%}, respectively. SelEx+SCAN also exceeds SEAL on several reported metrics without extra fine-grained supervision, although the comparison is auxiliary because SEAL uses hierarchical labels. The observed regressions are concentrated in particular host-baseline and subset combinations and may partly reflect keeping the evidence-slot hyperparameter $E$ fixed without per-domain tuning. Overall, the results are consistent with normal-anatomy suppression making subtle abnormalities more salient.

\paragraph{\textbf{Results on Within-Taxonomy Discovery.}}
\Cref{tab:setting3} verifies SCAN in within-taxonomy discovery, where new sibling subtypes differ by subtle, localized cues under a shared parent appearance. With SelEx, SCAN raises the average \textit{All} accuracy from 53.7\% to 66.0\% (+12.3\%), peaking on MLL23 (\textit{All} 62.5\% to 81.4\%, \textbf{+18.9\%}) and Gastrovision (\textit{Old} +\textbf{33.8\%}). SelEx+SCAN also exceeds SEAL on MLL23 \textit{New} accuracy (61.9\% vs. 47.2\%) without extra fine-grained supervision, although SEAL remains an auxiliary reference because it uses hierarchical labels. The observed drops are concentrated in particular datasets and host baselines and may partly reflect fixing $E$ across domains to stress robustness; overall, the results support our motivation that predictive suppression and surprise salience expose fine-grained subtype evidence.





\subsection{Mechanism and Qualitative Analysis}
\label{subsec:exp_Analysis}

\begin{figure}[tb]
  \centering
  \begin{subfigure}{0.24\linewidth}
    {\includegraphics[width=0.95\textwidth]{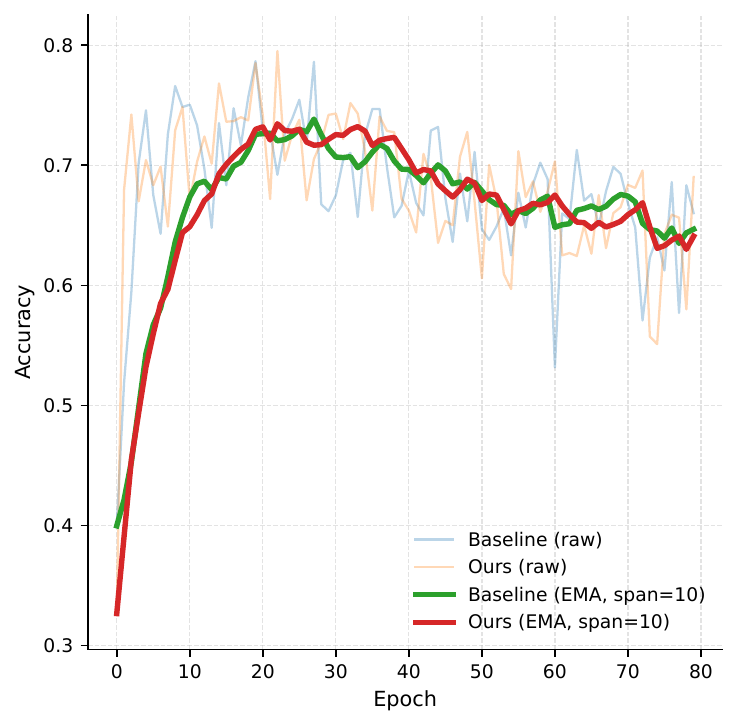}}
    \caption{Known Acc.}
    \label{fig:short-a}
  \end{subfigure}
  \hfill
  \begin{subfigure}{0.24\linewidth}
    {\includegraphics[width=0.95\textwidth]{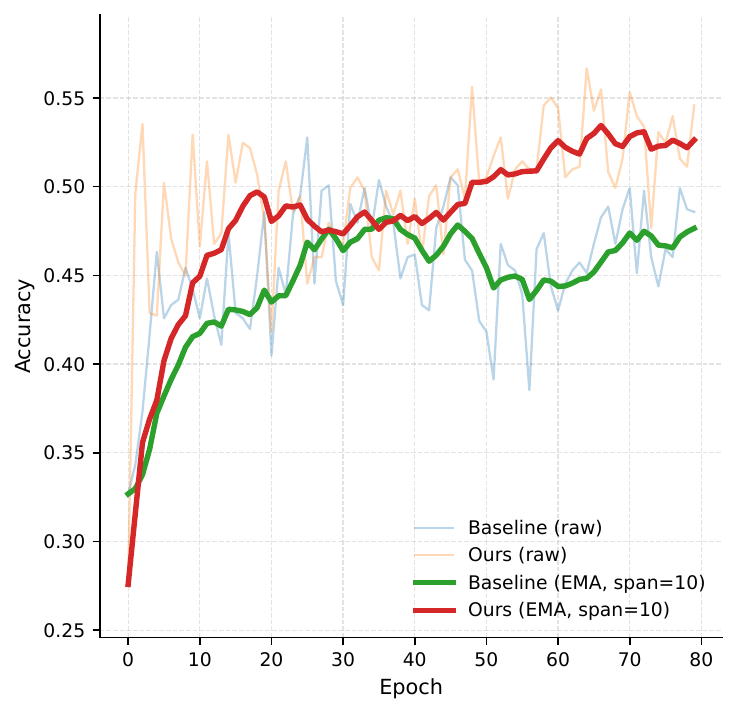}}
    \caption{Novel Acc.}
    \label{fig:short-b}
  \end{subfigure}
  \hfill
  \begin{subfigure}{0.24\linewidth}
{\includegraphics[width=0.95\textwidth]{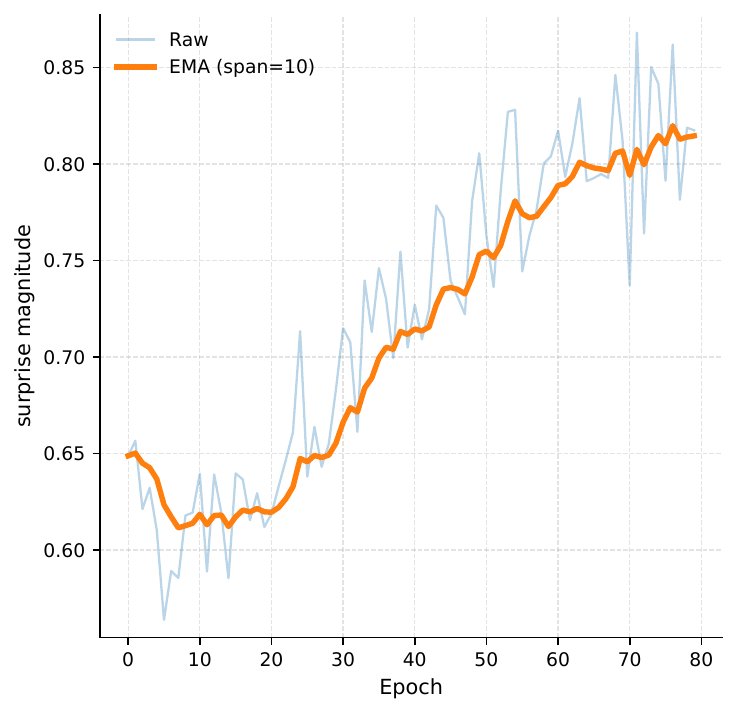}}
    \caption{$S$ in~\cref{eq:gate_terms}}
    \label{fig:short-c}
  \end{subfigure}
  \hfill
  \begin{subfigure}{0.24\linewidth}
    {\includegraphics[width=0.95\textwidth]{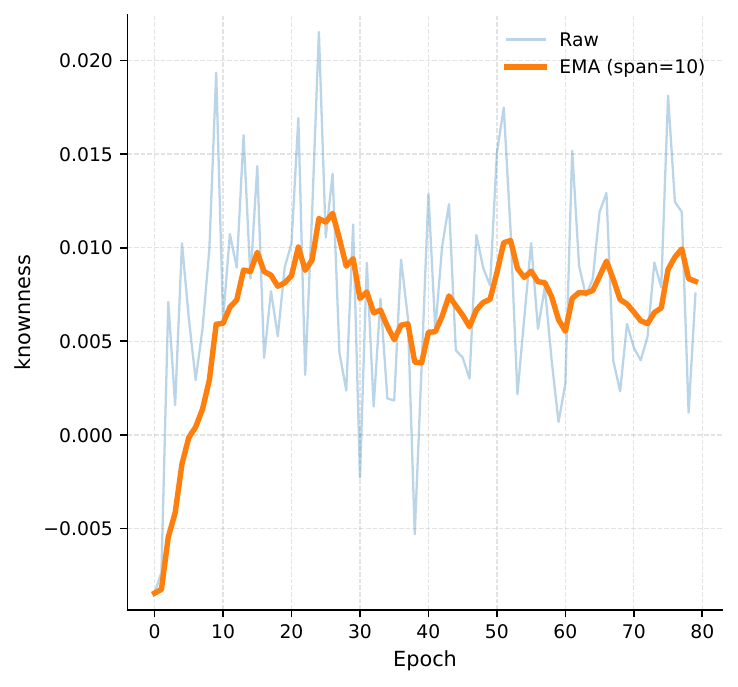}}
    \caption{$k$ in~\cref{eq:gate_terms}}
    \label{fig:short-d}
  \end{subfigure}
  \caption{Training dynamics on Gastrovision (Setting I): Known/Novel accuracy and the evolution of the surprise signal ($S$) and knownness ($k$) for SelEx and SelEx+SCAN.}
  \label{fig:training_dynamics}
\end{figure}

\paragraph{\textbf{Adaptive Discovery Dynamics.}}
We analyze Gastrovision (Setting I) training dynamics to understand SCAN's balance of discovery. As shown in~\cref{fig:training_dynamics}, SCAN keeps \emph{old} accuracy comparable to SelEx, while \emph{new} accuracy rises steadily and substantially exceeds the baseline. This improvement correlates with rising surprise signals over epochs, a pattern consistent with increased sensitivity to sparse, high-energy deviations. Around 40 epochs, SelEx shows a clear performance drop, whereas SCAN enters a second growth phase with fast accuracy gains. Meanwhile, the knownness indicator $k$ in~\cref{eq:gate_terms} slightly drops, corresponding to a less restrictive surprise gate for new samples. These trends suggest that SCAN preserves stable knowledge while reallocating representation capacity as training evolves.

\begin{figure}[t]
  \centering
  \begin{subfigure}[t]{0.32\linewidth}
    \centering
    \begin{minipage}[c][2.5cm][c]{\linewidth}
      \centering
      \includegraphics[width=\linewidth,height=2.35cm,keepaspectratio]{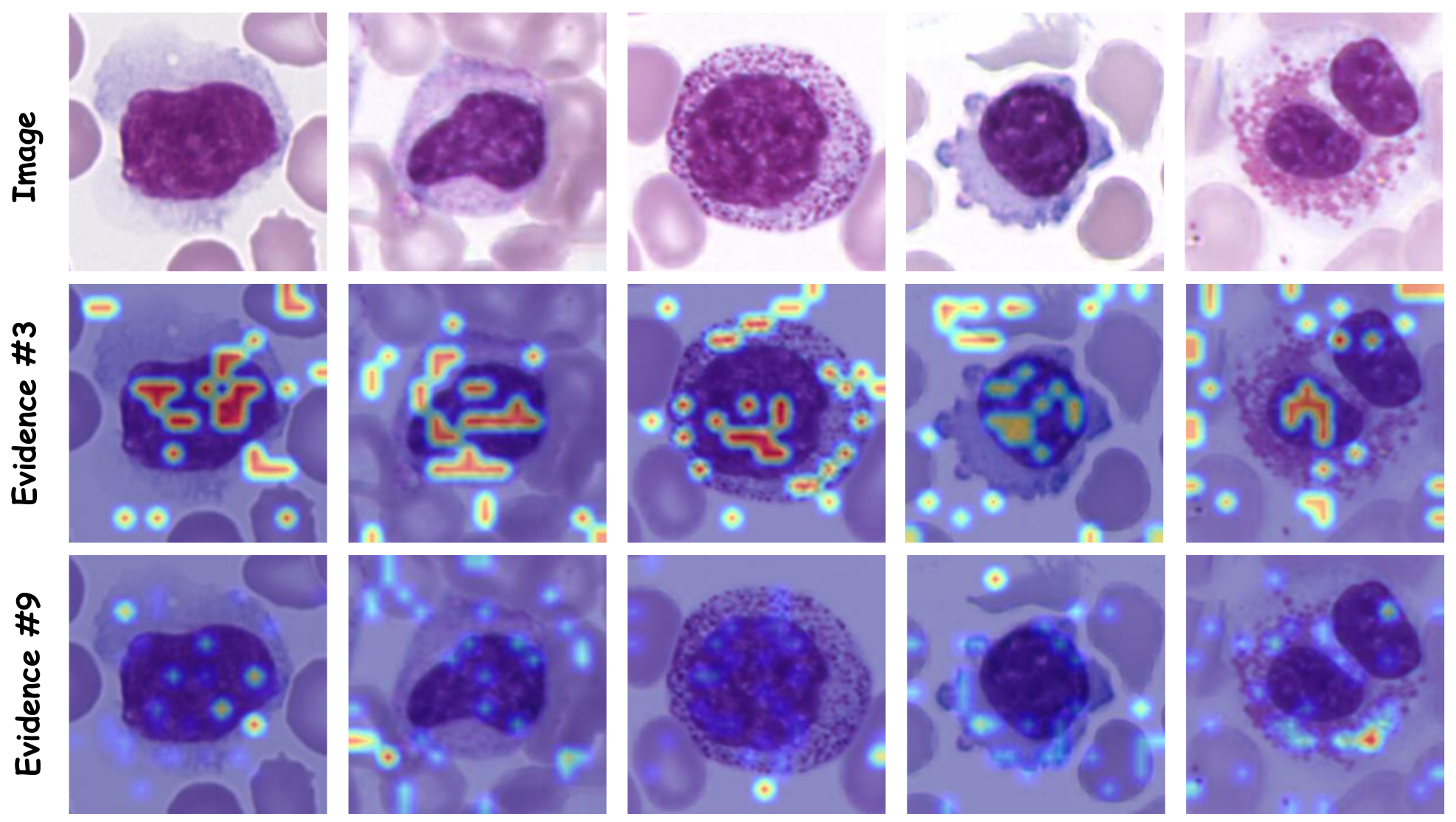}
    \end{minipage}
    \caption{Evidence activations.}
    \label{fig:evidence}
  \end{subfigure}
  \hfill
  \begin{subfigure}[t]{0.32\linewidth}
    \centering
    \begin{minipage}[c][2.5cm][c]{\linewidth}
      \centering
      \includegraphics[width=\linewidth,height=2.35cm,keepaspectratio]{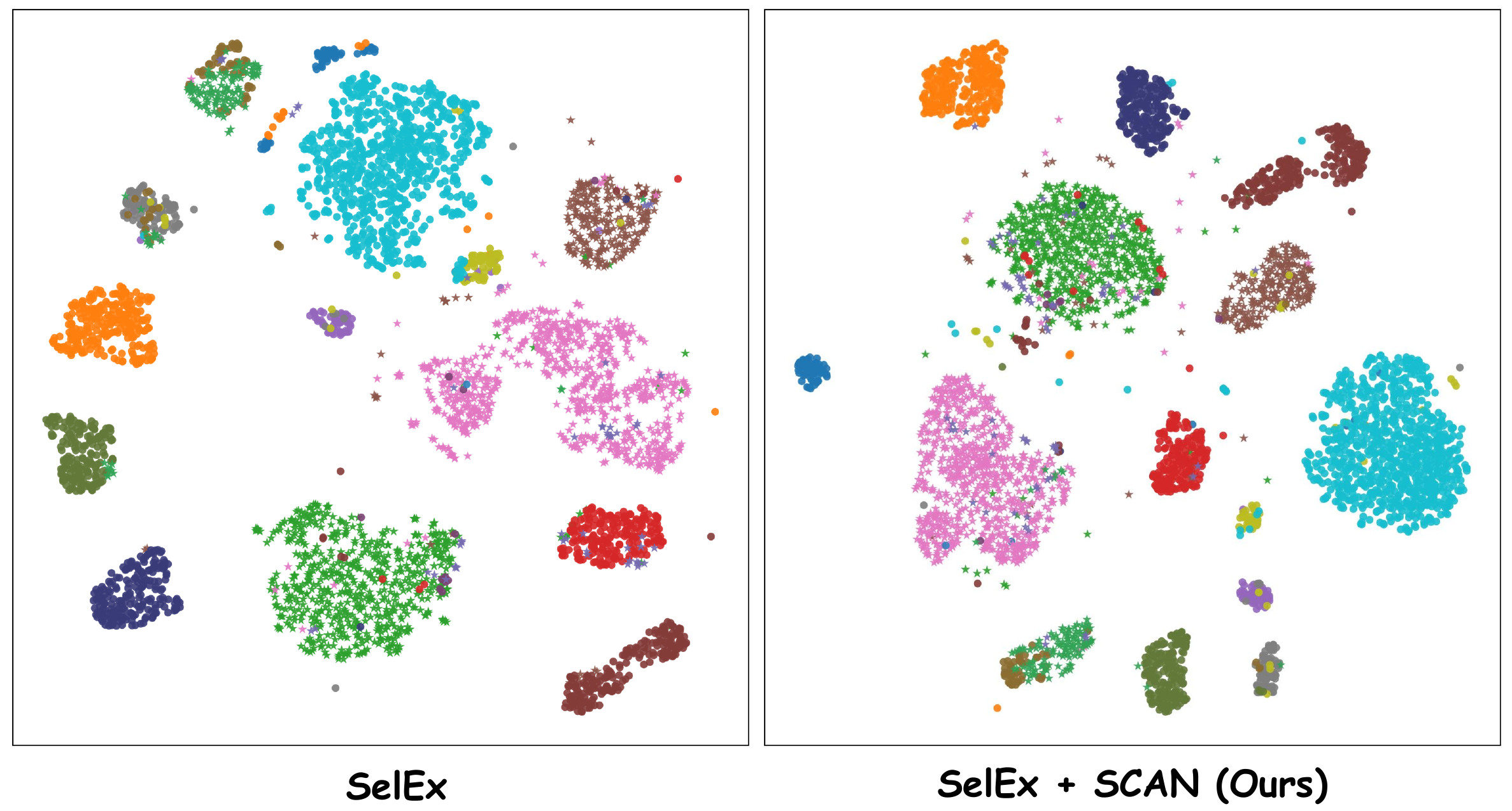}
    \end{minipage}
    \caption{UMAP embeddings.}
    \label{fig:umap}
  \end{subfigure}
  \hfill
  \begin{subfigure}[t]{0.32\linewidth}
    \centering
    \begin{minipage}[c][2.5cm][c]{\linewidth}
      \centering
      \renewcommand{\arraystretch}{0.85}
      \resizebox{\linewidth}{!}{%
        \begin{tabular}{cccc}
          \toprule
          \shortstack{Models\\ \ } & \shortstack{Params\\(M)} & \shortstack{Training\\Time (s)} & \shortstack{Inference\\Time (s)} \\
          \midrule
          SimGCD & 92.10 & 27.87 & 10.61 \\
          + SCAN & 99.68 & 30.12 & 11.32 \\
          \rowcolor{gray!15}$\Delta$ & + 7.6 & + 8\% & + 6\% \\
          \bottomrule
        \end{tabular}%
      }
    \end{minipage}
    \caption{Computational overhead.}
    \label{fig:computation_cost}
  \end{subfigure}
  \caption{Mechanism, qualitative, and efficiency analyses of SCAN.}
  \label{fig:analysis_summary}
\end{figure}

\paragraph{\textbf{Cross-Sample Consistency of Evidence Slots.}}
To examine the semantic consistency of learned evidence slots in~\cref{sec:predictive_suppression}, we visualize spatial activations on peripheral blood smear images. As shown in~\cref{fig:evidence}, the slots appear to capture recurring cellular patterns. Specifically, Evidence slot $\#3$ often highlights high-contrast punctate textures around the perinuclear cytoplasm and salient cell boundaries, whereas Evidence slot $\#9$ tends to cover diffuse, low-intensity staining patterns in broader nuclear or cytoplasmic regions. These qualitative patterns suggest that the learned slots may provide reusable morphological primitives across samples.

\paragraph{\textbf{Qualitative Analysis.}}
\Cref{fig:umap} visualizes embeddings on MLL23 (Setting I) with UMAP~\cite{umap}. SelEx yields dispersed clusters and noticeable overlap between new (stars) and old (dots) regions. With SCAN, the visualization shows tighter within-class neighborhoods and clearer margins among classes. New cells appear more concentrated into coherent groups and more separated from old clusters, suggesting improved organization of subtle tail concepts in this qualitative example.

\newpage

\paragraph{\textbf{Computational Overhead.}}
As shown in~\cref{fig:computation_cost}, SCAN adds a small number of parameters and several linear layers on top of the backbone and head. The measured increases in training and inference time are limited, suggesting that SCAN can be integrated into existing discovery pipelines with modest computational overhead.

\subsection{Discussion}

GBD is not intended as a leaderboard against text-supervised multimodal models (\emph{e.g.}, CLIP or LLMs), which assume a stable language anchor and reliable textual supervision. Instead, it targets upstream clinical \emph{concept formation}: organizing unlabeled archives into coherent cohorts that experts can verify, name, merge, or split before integration into guidelines and curated datasets. This matters because emerging diseases and subtypes often lack consistent nomenclature across sites and time, while clinicians require cohort-level discovery, representative prototypes, and similar-case lists rather than one-off predictions.

\section{Conclusion}

 We study generalized discovery in biomedical imaging under realistic clinical shifts. To support rigorous and comparable evaluation, we establish \textbf{Generalized Biomedicine Discovery}, a unified benchmark covering (i) long-tail rare disease discovery, (ii) normal-to-abnormal discovery, and (iii) within-taxonomy discovery across endoscopy, pathology, microscopy, and dermatology. Motivated by our central insight that dominant known patterns, especially normal anatomy, form a visual manifold that suppresses weak novelty cues, we propose \textbf{SCAN}, a cognition-inspired three-stage process of predictive suppression, surprise-evoked salience, and complementary accommodation. Extensive analyses and experiments support this insight, showing overall improvements in novel-disease discovery while revealing some dataset- and baseline-specific trade-offs in preserving established knowledge. GBD provides a clinically grounded, taxonomy-aware testbed that reframes biomedical imaging as a measurable concept-formation problem, enabling discovery systems to surface rare and emerging conditions and updateable subtypes with minimal reliance on exhaustive relabeling.

 \section*{Acknowledgments}
The work described in this paper is supported by grants from HKU Startup Fund and HKU Seed Fund for Basic Research.




%
%
\bibliographystyle{splncs04}
\bibliography{main}

@String(CVPR= {IEEE Conf. Comput. Vis. Pattern Recog.})

@String(ICCV= {Int. Conf. Comput. Vis.})

@String(ECCV= {Eur. Conf. Comput. Vis.})

@String(TOG= {ACM Trans. Graph.})

@String(ICLR = {Int. Conf. Learn. Represent.})

@String(AAAI = {AAAI})

@String(CVPR  = {CVPR})

@String(ICCV  = {ICCV})

@String(ECCV  = {ECCV})

@String(TOG   = {ACM TOG})

@String(ICLR  = {ICLR})

@String(ICML = {ICML})

@String(Arxiv = {Arxiv})

@inproceedings{ugcd,
  title={No representation rules them all in category discovery},
  author={Vaze, Sagar and Vedaldi, Andrea and Zisserman, Andrew},
  booktitle={NeurIPS},
  volume={36},
  year={2024}
}

@inproceedings{openmix,
  title={Openmix: Reviving known knowledge for discovering novel visual categories in an open world},
  author={Zhong, Zhun and Zhu, Linchao and Luo, Zhiming and Li, Shaozi and Yang, Yi and Sebe, Nicu},
  booktitle={CVPR},
  pages={9462--9470},
  year={2021}
}

@inproceedings{incd,
  title={Class-incremental novel class discovery},
  author={Roy, Subhankar and Liu, Mingxuan and Zhong, Zhun and Sebe, Nicu and Ricci, Elisa},
  booktitle={ECCV},
  pages={317--333},
  year={2022},
  organization={Springer}
}

@inproceedings{ncd,
  title={Learning to discover novel visual categories via deep transfer clustering},
  author={Han, Kai and Vedaldi, Andrea and Zisserman, Andrew},
  booktitle={ICCV},
  pages={8401--8409},
  year={2019}
}

@inproceedings{gcd,
  title={Generalized category discovery},
  author={Vaze, Sagar and Han, Kai and Vedaldi, Andrea and Zisserman, Andrew},
  booktitle=CVPR,
  pages={7492--7501},
  year={2022}
}

@inproceedings{gpc,
  title={Learning semi-supervised gaussian mixture models for generalized category discovery},
  author={Zhao, Bingchen and Wen, Xin and Han, Kai},
  booktitle={ICCV},
  pages={16623--16633},
  year={2023}
}

@inproceedings{dccl,
  title={Dynamic conceptional contrastive learning for generalized category discovery},
  author={Pu, Nan and Zhong, Zhun and Sebe, Nicu},
  booktitle={CVPR},
  pages={7579--7588},
  year={2023}
}

@inproceedings{simgcd,
  title={Parametric classification for generalized category discovery: A baseline study},
  author={Wen, Xin and Zhao, Bingchen and Qi, Xiaojuan},
  booktitle={ICCV},
  pages={16590--16600},
  year={2023}
}

@inproceedings{cms,
  title={Contrastive Mean-Shift Learning for Generalized Category Discovery},
  author={Choi, Sua and Kang, Dahyun and Cho, Minsu},
  booktitle={CVPR},
  pages={23094--23104},
  year={2024}
}

@inproceedings{pim,
  title={Parametric information maximization for generalized category discovery},
  author={Chiaroni, Florent and Dolz, Jose and Masud, Ziko Imtiaz and Mitiche, Amar and Ben Ayed, Ismail},
  booktitle={ICCV},
  pages={1729--1739},
  year={2023}
}

@inproceedings{sptnet,
  title={SPTNet: An efficient alternative framework for generalized category discovery with spatial prompt tuning},
  author={Wang, Hongjun and Vaze, Sagar and Han, Kai},
  booktitle={ICLR},
  year={2024}
}

@inproceedings{legogcd,
  title={Solving the catastrophic forgetting problem in generalized category discovery},
  author={Cao, Xinzi and Zheng, Xiawu and Wang, Guanhong and Yu, Weijiang and Shen, Yunhang and Li, Ke and Lu, Yutong and Tian, Yonghong},
  booktitle={CVPR},
  pages={16880--16889},
  year={2024}
}

@inproceedings{he2020momentum,
  title={Momentum contrast for unsupervised visual representation learning},
  author={He, Kaiming and Fan, Haoqi and Wu, Yuxin and Xie, Saining and Girshick, Ross},
  booktitle={CVPR},
  pages={9729--9738},
  year={2020}
}

@inproceedings{dino,
  title={Emerging properties in self-supervised vision transformers},
  author={Caron, Mathilde and Touvron, Hugo and Misra, Ishan and J{\'e}gou, Herv{\'e} and Mairal, Julien and Bojanowski, Piotr and Joulin, Armand},
  booktitle={ICCV},
  pages={9650--9660},
  year={2021}
}

@inproceedings{simclr,
  title={A simple framework for contrastive learning of visual representations},
  author={Chen, Ting and Kornblith, Simon and Norouzi, Mohammad and Hinton, Geoffrey},
  booktitle={ICML},
  pages={1597--1607},
  year={2020},
  organization={PmLR}
}

@inproceedings{ssb,
  author       = {Sagar Vaze and
                  Kai Han and
                  Andrea Vedaldi and
                  Andrew Zisserman},
  title        = {Open-Set Recognition: {A} Good Closed-Set Classifier is All You Need},
  booktitle    = {ICLR2022},
  publisher    = {OpenReview.net},
  year         = {2022},
}

@article{selex,
  title={SelEx: Self-Expertise in Fine-Grained Generalized Category Discovery},
  author={Rastegar, Sarah and Salehi, Mohammadreza and Asano, Yuki M and Doughty, Hazel and Snoek, Cees GM},
  journal={arXiv preprint arXiv:2408.14371},
  year={2024}
}

@article{chen1982topological,
  title={Topological structure in visual perception},
  author={Chen, Lin},
  journal={Science},
  volume={218},
  number={4573},
  pages={699--700},
  year={1982},
  publisher={American Association for the Advancement of Science}
}

@inproceedings{tang2025ocrt,
  title={OCRT: Boosting Foundation Models in the Open World with Object-Concept-Relation Triad},
  author={Tang, Luyao and Yuan, Yuxuan and Chen, Chaoqi and Zhang, Zeyu and Huang, Yue and Zhang, Kun},
  booktitle={Proceedings of the Computer Vision and Pattern Recognition Conference},
  pages={25422--25433},
  year={2025}
}

@inproceedings{tang2025dissecting,
  title={Dissecting generalized category discovery: Multiplex consensus under self-deconstruction},
  author={Tang, Luyao and Huang, Kunze and Chen, Chaoqi and Yuan, Yuxuan and Li, Chenxin and Tu, Xiaotong and Ding, Xinghao and Huang, Yue},
  booktitle={Proceedings of the IEEE/CVF International Conference on Computer Vision},
  pages={297--307},
  year={2025}
}

@article{geng2020recent,
  title={Recent advances in open set recognition: A survey},
  author={Geng, Chuanxing and Huang, Sheng-jun and Chen, Songcan},
  journal={IEEE transactions on pattern analysis and machine intelligence},
  volume={43},
  number={10},
  pages={3614--3631},
  year={2020},
  publisher={IEEE}
}

@article{scheirer2012toward,
  title={Toward open set recognition},
  author={Scheirer, Walter J and de Rezende Rocha, Anderson and Sapkota, Archana and Boult, Terrance E},
  journal={IEEE transactions on pattern analysis and machine intelligence},
  volume={35},
  number={7},
  pages={1757--1772},
  year={2012},
  publisher={IEEE}
}

@incollection{thagard2013cognitive,
  title={Cognitive science},
  author={Thagard, Paul},
  booktitle={The Routledge companion to philosophy of science},
  pages={597--608},
  year={2013},
  publisher={Routledge}
}

@article{blasi2022over,
  title={Over-reliance on English hinders cognitive science},
  author={Blasi, Dami{\'a}n E and Henrich, Joseph and Adamou, Evangelia and Kemmerer, David and Majid, Asifa},
  journal={Trends in cognitive sciences},
  volume={26},
  number={12},
  pages={1153--1170},
  year={2022},
  publisher={Elsevier}
}

@inproceedings{yang2024language,
  title={Language models as inductive reasoners},
  author={Yang, Zonglin and Dong, Li and Du, Xinya and Cheng, Hao and Cambria, Erik and Liu, Xiaodong and Gao, Jianfeng and Wei, Furu},
  booktitle={Proceedings of the 18th Conference of the European Chapter of the Association for Computational Linguistics (Volume 1: Long Papers)},
  pages={209--225},
  year={2024}
}

@article{sung2015data,
  title={Data-driven structural priors for shape completion},
  author={Sung, Minhyuk and Kim, Vladimir G and Angst, Roland and Guibas, Leonidas},
  journal={ACM Transactions on Graphics (TOG)},
  volume={34},
  number={6},
  pages={1--11},
  year={2015},
  publisher={ACM New York, NY, USA}
}

@article{wen2022mvs,
  title={MVS-GCN: A prior brain structure learning-guided multi-view graph convolution network for autism spectrum disorder diagnosis},
  author={Wen, Guangqi and Cao, Peng and Bao, Huiwen and Yang, Wenju and Zheng, Tong and Zaiane, Osmar},
  journal={Computers in biology and medicine},
  volume={142},
  pages={105239},
  year={2022},
  publisher={Elsevier}
}

@inproceedings{roessle2022dense,
  title={Dense depth priors for neural radiance fields from sparse input views},
  author={Roessle, Barbara and Barron, Jonathan T and Mildenhall, Ben and Srinivasan, Pratul P and Nie{\ss}ner, Matthias},
  booktitle={Proceedings of the IEEE/CVF conference on computer vision and pattern recognition},
  pages={12892--12901},
  year={2022}
}

@article{song2023flow,
  title={Flow factorized representation learning},
  author={Song, Yue and Keller, Andy and Sebe, Nicu and Welling, Max},
  journal={Advances in Neural Information Processing Systems},
  volume={36},
  pages={49761--49782},
  year={2023}
}

@article{liang2023factorized,
  title={Factorized contrastive learning: Going beyond multi-view redundancy},
  author={Liang, Paul Pu and Deng, Zihao and Ma, Martin Q and Zou, James Y and Morency, Louis-Philippe and Salakhutdinov, Ruslan},
  journal={Advances in Neural Information Processing Systems},
  volume={36},
  pages={32971--32998},
  year={2023}
}

@inproceedings{muennighoff2023crosslingual,
  title={Crosslingual generalization through multitask finetuning},
  author={Muennighoff, Niklas and Wang, Thomas and Sutawika, Lintang and Roberts, Adam and Biderman, Stella and Le Scao, Teven and Bari, M Saiful and Shen, Sheng and Yong, Zheng-Xin and Schoelkopf, Hailey and others},
  booktitle={Proceedings of the 61st Annual Meeting of the Association for Computational Linguistics (Volume 1: Long Papers)},
  pages={15991--16111},
  year={2023}
}

@inproceedings{zhang2023federated,
  title={Federated domain generalization with generalization adjustment},
  author={Zhang, Ruipeng and Xu, Qinwei and Yao, Jiangchao and Zhang, Ya and Tian, Qi and Wang, Yanfeng},
  booktitle={Proceedings of the IEEE/CVF Conference on Computer Vision and Pattern Recognition},
  pages={3954--3963},
  year={2023}
}

@article{saito2021openmatch,
  title={Openmatch: Open-set semi-supervised learning with open-set consistency regularization},
  author={Saito, Kuniaki and Kim, Donghyun and Saenko, Kate},
  journal={Advances in Neural Information Processing Systems},
  volume={34},
  pages={25956--25967},
  year={2021}
}

@inproceedings{yoshihashi2019classification,
  title={Classification-reconstruction learning for open-set recognition},
  author={Yoshihashi, Ryota and Shao, Wen and Kawakami, Rei and You, Shaodi and Iida, Makoto and Naemura, Takeshi},
  booktitle={Proceedings of the IEEE/CVF conference on computer vision and pattern recognition},
  pages={4016--4025},
  year={2019}
}

@article{zhang2025generalist,
  title={A generalist foundation model and database for open-world medical image segmentation},
  author={Zhang, Siqi and Zhang, Qizhe and Zhang, Shanghang and Liu, Xiaohong and Yue, Jingkun and Lu, Ming and Xu, Huihuan and Yao, Jiaxin and Wei, Xiaobao and Cao, Jiajun and others},
  journal={Nature Biomedical Engineering},
  pages={1--16},
  year={2025},
  publisher={Nature Publishing Group UK London}
}

@inproceedings{yang2025medical,
  title={Medical World Model},
  author={Yang, Yijun and Wang, Zhao-Yang and Liu, Qiuping and Sun, Shuwen and Wang, Kang and Chellappa, Rama and Zhou, Zongwei and Yuille, Alan and Zhu, Lei and Zhang, Yu-Dong and others},
  booktitle={Proceedings of the IEEE/CVF International Conference on Computer Vision},
  pages={8319--8329},
  year={2025}
}

@article{yang2024limits,
  title={The limits of fair medical imaging AI in real-world generalization},
  author={Yang, Yuzhe and Zhang, Haoran and Gichoya, Judy W and Katabi, Dina and Ghassemi, Marzyeh},
  journal={Nature medicine},
  volume={30},
  number={10},
  pages={2838--2848},
  year={2024},
  publisher={Nature Publishing Group US New York}
}

@inproceedings{yang2022proco,
  title={Proco: Prototype-aware contrastive learning for long-tailed medical image classification},
  author={Yang, Zhixiong and Pan, Junwen and Yang, Yanzhan and Shi, Xiaozhou and Zhou, Hong-Yu and Zhang, Zhicheng and Bian, Cheng},
  booktitle={International conference on medical image computing and computer-assisted intervention},
  pages={173--182},
  year={2022},
  organization={Springer}
}

@article{zhang2023deep,
  title={Deep long-tailed learning: A survey},
  author={Zhang, Yifan and Kang, Bingyi and Hooi, Bryan and Yan, Shuicheng and Feng, Jiashi},
  journal={IEEE transactions on pattern analysis and machine intelligence},
  volume={45},
  number={9},
  pages={10795--10816},
  year={2023},
  publisher={IEEE}
}

@article{zheng2024large,
  title={Large-scale long-tailed disease diagnosis on radiology images},
  author={Zheng, Qiaoyu and Zhao, Weike and Wu, Chaoyi and Zhang, Xiaoman and Dai, Lisong and Guan, Hengyu and Li, Yuehua and Zhang, Ya and Wang, Yanfeng and Xie, Weidi},
  journal={Nature Communications},
  volume={15},
  number={1},
  pages={10147},
  year={2024},
  publisher={Nature Publishing Group UK London}
}

@inproceedings{holste2022long,
  title={Long-tailed classification of thorax diseases on chest x-ray: A new benchmark study},
  author={Holste, Gregory and Wang, Song and Jiang, Ziyu and Shen, Thomas C and Shih, George and Summers, Ronald M and Peng, Yifan and Wang, Zhangyang},
  booktitle={MICCAI Workshop on Data Augmentation, Labelling, and Imperfections},
  pages={22--32},
  year={2022},
  organization={Springer}
}

@article{thirion2000statistical,
  title={Statistical analysis of normal and abnormal dissymmetry in volumetric medical images},
  author={Thirion, J-P and Prima, Sylvain and Subsol, G{\'e}rard and Roberts, Neil},
  journal={Medical Image Analysis},
  volume={4},
  number={2},
  pages={111--121},
  year={2000},
  publisher={Elsevier}
}

@book{bushberg2011essential,
  title={The essential physics of medical imaging},
  author={Bushberg, Jerrold T and Boone, John M},
  year={2011},
  publisher={Lippincott Williams \& Wilkins}
}

@book{feeman2010mathematics,
  title={The mathematics of medical imaging},
  author={Feeman, Timothy G},
  year={2010},
  publisher={Springer}
}

@article{dimitrovski2011hierarchical,
  title={Hierarchical annotation of medical images},
  author={Dimitrovski, Ivica and Kocev, Dragi and Loskovska, Suzana and D{\v{z}}eroski, Sa{\v{s}}o},
  journal={Pattern Recognition},
  volume={44},
  number={10-11},
  pages={2436--2449},
  year={2011},
  publisher={Elsevier}
}

@article{chen2020deep,
  title={Deep hiearchical multi-label classification applied to chest X-ray abnormality taxonomies},
  author={Chen, Haomin and Miao, Shun and Xu, Daguang and Hager, Gregory D and Harrison, Adam P},
  journal={Medical image analysis},
  volume={66},
  pages={101811},
  year={2020},
  publisher={Elsevier}
}

@article{yu2025hierarchical,
  title={Hierarchical skin lesion image classification with prototypical decision tree},
  author={Yu, Zhen and Nguyen, Toan D and Ju, Lie and Gal, Yaniv and Sashindranath, Maithili and Bonnington, Paul and Zhang, Lei and Mar, Victoria and Ge, Zongyuan},
  journal={NPJ Digital Medicine},
  volume={8},
  number={1},
  pages={26},
  year={2025},
  publisher={Nature Publishing Group UK London}
}

@article{lester1999survey,
  title={A survey of hierarchical non-linear medical image registration},
  author={Lester, Hava and Arridge, Simon R},
  journal={Pattern recognition},
  volume={32},
  number={1},
  pages={129--149},
  year={1999},
  publisher={Elsevier}
}

@article{yao2023adversarial,
  title={Adversarial medical image with hierarchical feature hiding},
  author={Yao, Qingsong and He, Zecheng and Li, Yuexiang and Lin, Yi and Ma, Kai and Zheng, Yefeng and Zhou, S Kevin},
  journal={IEEE Transactions on Medical Imaging},
  volume={43},
  number={4},
  pages={1296--1307},
  year={2023},
  publisher={IEEE}
}

@article{feng2025neighbor,
  title={Neighbor-Guided Unbiased Framework for Generalized Category Discovery in Medical Image Classification},
  author={Feng, Wei and Zhou, Sijin and Jiang, Yiwen and Tang, Feilong and Ge, Zongyuan},
  journal={IEEE Journal of Biomedical and Health Informatics},
  year={2025},
  publisher={IEEE}
}

@inproceedings{das2025medgcd,
  title={MedGCD: Generalized Category Discovery in Medical Imaging},
  author={Das, Ankit and Gautam, Chandan and Agrawal, Pritee and Yang, Feng and Liu, Yong and Savitha, Ramasamy},
  booktitle={International Conference on Medical Image Computing and Computer-Assisted Intervention},
  pages={427--437},
  year={2025},
  organization={Springer}
}

@article{he2025category,
  title={Category discovery: An open-world perspective},
  author={He, Zhenqi and Liu, Yuanpei and Han, Kai},
  journal={arXiv preprint arXiv:2509.22542},
  year={2025}
}

@article{abu2017medical,
  title={Medical expert systems survey},
  author={Abu-Nasser, Bassem S},
  journal={International Journal of Engineering and Information Systems},
  volume={1},
  number={7},
  pages={218--224},
  year={2017},
  publisher={MISC}
}

@article{shortliffe1986medical,
  title={Medical expert systems—knowledge tools for physicians},
  author={Shortliffe, Edward H},
  journal={Western Journal of Medicine},
  volume={145},
  number={6},
  pages={830},
  year={1986}
}

@article{gilhooly1990cognitive,
  title={Cognitive psychology and medical diagnosis},
  author={Gilhooly, Kenneth J},
  journal={Applied cognitive psychology},
  volume={4},
  number={4},
  pages={261--272},
  year={1990},
  publisher={Wiley Online Library}
}

@article{van2013cognitive,
  title={Cognitive diagnostic error in internal medicine},
  author={Van den Berge, Kees and Mamede, S{\'\i}lvia},
  journal={European journal of internal medicine},
  volume={24},
  number={6},
  pages={525--529},
  year={2013},
  publisher={Elsevier}
}

@article{croskerry2023challenge,
  title={The challenge of cognitive science for medical diagnosis},
  author={Croskerry, Pat and Campbell, Samuel G and Petrie, David A},
  journal={Cognitive Research: Principles and Implications},
  volume={8},
  number={1},
  pages={13},
  year={2023},
  publisher={Springer}
}

@article{rao1999predictive,
  title={Predictive coding in the visual cortex: a functional interpretation of some extra-classical receptive-field effects},
  author={Rao, Rajesh PN and Ballard, Dana H},
  journal={Nature neuroscience},
  volume={2},
  number={1},
  pages={79--87},
  year={1999},
  publisher={Nature Publishing Group}
}

@article{itti2009bayesian,
  title={Bayesian surprise attracts human attention},
  author={Itti, Laurent and Baldi, Pierre},
  journal={Vision research},
  volume={49},
  number={10},
  pages={1295--1306},
  year={2009},
  publisher={Elsevier}
}

@article{mcclelland1995there,
  title={Why there are complementary learning systems in the hippocampus and neocortex: insights from the successes and failures of connectionist models of learning and memory.},
  author={McClelland, James L and McNaughton, Bruce L and O'Reilly, Randall C},
  journal={Psychological review},
  volume={102},
  number={3},
  pages={419},
  year={1995},
  publisher={American Psychological Association}
}

@article{oquab2023dinov2,
  title={Dinov2: Learning robust visual features without supervision},
  author={Oquab, Maxime and Darcet, Timoth{\'e}e and Moutakanni, Th{\'e}o and Vo, Huy and Szafraniec, Marc and Khalidov, Vasil and Fernandez, Pierre and Haziza, Daniel and Massa, Francisco and El-Nouby, Alaaeldin and others},
  journal={arXiv preprint arXiv:2304.07193},
  year={2023}
}

@inproceedings{he2025seal,
  title={Seal: Semantic-aware hierarchical learning for generalized category discovery},
  author={He, Zhenqi and Liu, Yuanpei and Han, Kai},
  booktitle={The Thirty-ninth Annual Conference on Neural Information Processing Systems},
  year={2025}
}

@inproceedings{jha2023gastrovision,
  title={GastroVision: A Multi-class Endoscopy Image Dataset for Computer Aided Gastrointestinal Disease Detection},
  author={Jha, Debesh and Sharma, Vanshali and Dasu, Neethi and Tomar, Nikhil Kumar and Hicks, Steven and Bhuyan, M. K. and Das, Pradip K. and Riegler, Michael A. and Halvorsen, P{\aa}l and de Lange, Thomas and Bagci, Ulas},
  booktitle={ICML Workshop on Machine Learning for Multimodal Healthcare Data (ML4MHD 2023)},
  year={2023}
}

@misc{rafi2025histoset,
  author    = {Rafi, Abid Hasan and Bhowmik, Pankaj},
  title     = {{HistoSet-5$\times$14: A Collection of Balanced Multi-Organ Histopathology Datasets}},
  year      = {2025},
  publisher = {Mendeley Data},
  version   = {V1},
  doi       = {10.17632/nc8k63t7mp.1},
  url       = {https://doi.org/10.17632/nc8k63t7mp.1} 
}

@misc{mll23,
  author       = {Shetab Boushehri, Sayedali and
                  Gruber, Armin and
                  Sun, Xudong and
                  Kazeminia, Salome and
                  Matek, Christian and
                  Hehr, Matthias and
                  Spiekermann, Karsten and
                  Pohlkamp, Christian and
                  Haferlach, Torsten and
                  Marr, Carsten},
  title        = {A large publicly available single-cell peripheral
                   blood dataset (MLL23)
                  },
  month        = jan,
  year         = 2023,
  publisher    = {Zenodo},
  doi          = {10.5281/zenodo.14277609},
  url          = {https://doi.org/10.5281/zenodo.14277609},
}

@article{yilmaz2024derm12345,
  title={Derm12345: A large, multisource dermatoscopic skin lesion dataset with 40 subclasses},
  author={Yilmaz, Abdurrahim and Yasar, Sirin Pekcan and Gencoglan, Gulsum and Temelkuran, Burak},
  journal={Scientific Data},
  volume={11},
  number={1},
  pages={1302},
  year={2024},
  publisher={Nature Publishing Group UK London}
}

@article{umap,
  title={UMAP: Uniform Manifold Approximation and Projection},
  author={McInnes, Leland and Healy, John and Saul, Nathaniel and Grossberger, Lukas},
  journal={The Journal of Open Source Software},
  volume={3},
  number={29},
  pages={861},
  year={2018}
}

@inproceedings{he2026flowcomposer,
  title={FlowComposer: Composable Flows for Compositional Zero-Shot Learning},
  author={He, Zhenqi and Li, Lin and Chen, Long},
  booktitle={Proceedings of the IEEE/CVF Conference on Computer Vision and Pattern Recognition},
  pages={12396--12405},
  year={2026}
}

@inproceedings{liu2025hyperbolic,
  title={Hyperbolic category discovery},
  author={Liu, Yuanpei and He, Zhenqi and Han, Kai},
  booktitle={Proceedings of the Computer Vision and Pattern Recognition Conference},
  pages={9891--9900},
  year={2025}
}

@inproceedings{he2023transnuseg,
  title={Transnuseg: A lightweight multi-task transformer for nuclei segmentation},
  author={He, Zhenqi and Unberath, Mathias and Ke, Jing and Shen, Yiqing},
  booktitle={International Conference on Medical Image Computing and Computer-Assisted Intervention},
  pages={206--215},
  year={2023},
  organization={Springer}
}

@inproceedings{
liu2025debgcd,
title={Deb{GCD}: Debiased Learning with Distribution Guidance for Generalized Category Discovery},
author={Yuanpei Liu and Kai Han},
booktitle={The Thirteenth International Conference on Learning Representations},
year={2025},
url={https://openreview.net/forum?id=9B8o9AxSyb}
}

@inproceedings{fengprism,
  title={PRISM: Progressive Robust Learning for Open-World Continual Category Discovery},
  author={Feng, Wei and Zhou, Sijin and Jiang, Yiwen and Ge, Zongyuan},
  booktitle={The Fourteenth International Conference on Learning Representations},
  year={2026}
}

@article{feng2026leveraging,
  title={Leveraging Image-text Pairs for Generalized Category Discovery in Medical Image Classification},
  author={Feng, Wei and Wang, Bingjie and Wang, Zhonghua and Zhou, Sijin and Ge, Zongyuan},
  journal={IEEE Transactions on Medical Imaging},
  year={2026},
  publisher={IEEE}
}

@inproceedings{feng2026beyond,
  title={Beyond the Static World: Continual Category Discovery under Visual Drift},
  author={Feng, Wei and Jiang, Yiwen and Zhou, Sijin and Ge, Zongyuan},
  booktitle={Proceedings of the IEEE/CVF Conference on Computer Vision and Pattern Recognition},
  pages={25032--25042},
  year={2026}
}

@inproceedings{feng2026seeing,
  title={Seeing Through the Shift: Causality-Inspired Robust Generalized Category Discovery},
  author={Feng, Wei and Jiang, Yiwen and Zhou, Sijin and Qi, Zhuang and Xu, Zhongxing and Wang, Zhonghua and Tang, Feilong and Ge, Zongyuan},
  booktitle={Proceedings of the IEEE/CVF Conference on Computer Vision and Pattern Recognition},
  pages={17766--17775},
  year={2026}
}

@article{Bian2026Foundation, 
title={Foundation-Adaptive Integrated Refinement for Generalized Category Discovery},
volume={40}, 
number={4}, 
journal={AAAI}, 
author={Bian, Yuwei and Wang, Shidong and Yao, Yazhou and Zhang, Haofeng}, 
year={2026}, 
month={Mar.}, 
pages={2453-2461} 
}

@inproceedings{he2026harnessing,
  title={Harnessing textual semantic priors for knowledge transfer and refinement in clip-driven continual learning},
  author={He, Lingfeng and Cheng, De and Xu, Di and Wang, Huaijie and Wang, Nannan},
  booktitle={Proceedings of the AAAI Conference on Artificial Intelligence},
  volume={40},
  number={26},
  pages={21645--21653},
  year={2026}
}

@inproceedings{
he2026task,
title={Task-Driven Subspace Decomposition for Knowledge Sharing and Isolation in LoRA-based Continual Learning},
author={Lingfeng He and De Cheng and Huaijie Wang and Xiaofeng Zhu and Xi Yang and Nannan Wang and Xinbo Gao},
booktitle={Forty-third International Conference on Machine Learning},
year={2026}
}

@inproceedings{tangbures,
  title={Bures-Isotropy Alignment: Manifold Learning of Generalized Category Discovery},
  author={Tang, Luyao and Huang, Kunze and Chen, Chaoqi and Chen, Cheng},
  booktitle={The Fourteenth International Conference on Learning Representations}
}

@inproceedings{tangcoge,
  title={CoGe-GCD: Reframing Generalized Category Discovery with Compositional Generalization},
  author={Tang, Luyao and Zheng, Jiewei and Huang, Kunze and Chen, Chaoqi and Huang, Yue and Chen, Cheng},
  booktitle={Forty-third International Conference on Machine Learning}
}

@inproceedings{chen2023activate,
  title={Activate and reject: towards safe domain generalization under category shift},
  author={Chen, Chaoqi and Tang, Luyao and Tao, Leitian and Zhou, Hong-Yu and Huang, Yue and Han, Xiaoguang and Yu, Yizhou},
  booktitle={2023 IEEE/CVF International Conference on Computer Vision (ICCV)},
  pages={11518--11529},
  year={2023},
  organization={IEEE}
}

@article{chen2022mix,
  title={Mix and reason: Reasoning over semantic topology with data mixing for domain generalization},
  author={Chen, Chaoqi and Tang, Luyao and Liu, Feng and Zhao, Gangming and Huang, Yue and Yu, Yizhou},
  journal={Advances in Neural Information Processing Systems},
  volume={35},
  pages={33302--33315},
  year={2022}
}
\end{document}